\documentclass[review]{elsarticle}

\usepackage{lineno}
\modulolinenumbers[200]

\usepackage{amsfonts}
\usepackage{latexsym}
\usepackage{times}
\usepackage{amsmath}
\usepackage{amssymb}
\usepackage{caption}
\usepackage{graphicx}
\usepackage{url}
\usepackage{multirow}
\usepackage{color}

\journal{Arxiv}

\begin{document}

\begin{frontmatter}

\title{Knowledge Representation Learning: A Quantitative Review}

\author[mymainaddress]{Yankai Lin}
\ead{linyk14@mails.tsinghua.edu.cn}

\author[mymainaddress]{Xu Han}
\ead{thu.hanxu13@gmail.com}

\author[mymainaddress]{Ruobing Xie}
\ead{xrbsnowing@163.com}

\author[mymainaddress]{Zhiyuan Liu \corref{mycorrespondingauthor}}
\cortext[mycorrespondingauthor]{Corresponding author}
\ead{liuzy@tsinghua.edu.cn}

\author[mymainaddress]{Maosong Sun}
\ead{sms@tsinghua.edu.cn}

\address[mymainaddress]{FIT Building 4506, Tsinghua University, Beijing, China}

\begin{abstract}
Knowledge representation learning (KRL) aims to represent entities and relations in knowledge graph in low-dimensional semantic space, which have been widely used in massive knowledge-driven tasks. In this article, we introduce the reader to the motivations for KRL, and overview existing approaches for KRL. Afterwards, we extensively conduct and quantitative comparison and analysis of several typical KRL methods on three evaluation tasks of knowledge acquisition including knowledge graph completion, triple classification, and relation extraction. We also review the real-world applications of KRL, such as language modeling, question answering, information retrieval, and recommender systems. Finally, we discuss the remaining challenges and outlook the future directions for KRL. The codes and datasets  used in the experiments can be found in \url{https://github.com/thunlp/OpenKE}.
\end{abstract}


\end{frontmatter}

\linenumbers

\section{Introduction}
In recent years, people have built a large amount of knowledge graphs (KGs) such as Freebase~\cite{bollacker2008freebase}, DBpedia~\cite{auer2007dbpedia}, YAGO~\cite{suchanek2007yago}, NELL~\cite{carlson2010nell} and Wikidata~\cite{vrandevcic2014wikidata}. KGs provide us a novel aspect to describe the real world, which stores structured relational facts of concrete entities and abstract concepts in the real world. The structured relational facts could be either automatically extracted from enormous plaintexts and structured Web data, or manually annotated by human experts. To store these knowledge, KGs mainly contain two elements, i.e., entities that represent both concrete and abstract concepts, and relations that indicate relationships between entities. To record relational facts in KGs, many schemes such as RDF (resource description framework), have been proposed and typically represent those entities and relations in KGs as discrete symbols. For example, we know that Beijing is the capital of China. In KGs, we will represent this fact with the triple form as (\emph{Beijing}, \texttt{is\_capital\_of}, \emph{China}). Nowadays, these KGs play an important role in many tasks in artificial intelligence,  such as word similarity computation~\cite{pedersen2004wordnet}, word sense disambiguation~\cite{leacock1998combining,chen2014unified}, entity disambiguation~\cite{dredze2010entity}, semantic parsing~\cite{bordes2012joint,berant2013semantic}, text classification~\cite{scott1998text,wang2009using}, topic indexing~\cite{medelyan2008topic}, document summarization~\cite{verma2007semantic}, document ranking~\cite{hu2009understanding}, information extraction~\cite{hoffmann2011knowledge,daiber2013improving}, and question answering~\cite{bordes2014question,bordes2014open}.

    However, people are still facing two main challenges to utilize KGs in real-world application: data sparsity and growing computational inefficiency. Existing knowledge construction and application approaches \cite{lao2010relational,lao2012reading,di2012linked,smirnov2015patterns} usually store relation facts in KGs with one-hot representations of entities and relations which cannot afford their rich semantic information. One-hot representation, in essence, maps each entity or relation to an index, which can be very efficient for storage. However, it does not embed any semantic aspect of entities and relations. Hence, it cannot distinguish the similarities and differences among ``Bill Gates'',``Steve Jobs'' and ``United States''. Moreover, these works rely on designed sophisticated and specialized features extracted from external information sources or network structure of KGs. With the increasing the KG's size, these methods usually suffer from the issue of computational inefficiency and the lack of extensibility.


    
    

With the development of deep learning, distributed representation learning has shown their abilities in computer vision and natural language processing. Recently, distributed representation learning of KGs has also been explored, showing its powerful capability of representing knowledge in relation extraction, knowledge inference, and other knowledge-driven applications. Knowledge representation learning (KRL) typically learns the distributed representations of  both entities and relations of a KG, and projects their distributed representations into a low-dimensional semantic space. KRL usually wants to encode the semantic meaning of entities and relations with their corresponding low-dimensional vectors. Compared with the traditional representation, KRL gives the entities and relations in KG much dense representations, which leads to lower computational complexity in its applications. Moreover, KRL can explicitly capture the similarity between entities and relations via measuring the similarity of their low-dimensional embeddings. With the advantages above, KRL is blooming in the applications of KGs. Up till now, there are a great number of methods having been proposed using representation learning in KGs. 

In this article, we first review the recent advances in KRL. Second, we perform quantitative analysis of most existing KRL models on three typical tasks of knowledge acquisition including knowledge graph completion, triple classification, and relation extraction. Third, we introduce typical applications of KRL in real world such as recommendation system, language modeling, question answering, etc. Finally, we re-examine the remaining research challenges and outlook the trends for KRL and its applications.


\section{Knowledge Representation Learning}

Knowledge representation learning aims to embed the entities and relations in KGs into a low-dimensional continuous semantic space. For the convenience of presentation, we will introduce the basic notations used in this paper at the beginning. First, we define $G = (E, R, S)$ as a KG, where $E = \{e_1, e_2, \cdots, e_{|E|}\}$ is a set of $|E|$ entities, $R = \{r_1, r_2, \cdots\\, r_{|R|}\}$ is a set of $|R|$ relations, and $S \subseteq E\times R\times E$ is the set of fact triples with the  format $(h, r, t)$. Here, $h$ and $t$ indicate the head and tail entities, and $r$ indicates the relationship between them. For example, (\emph{Microsoft}, \texttt{founder}, \emph{Bill Gates}) indicates that there is a relation \texttt{founder} between \emph{Microsoft} and \emph{Bill Gates}.

Recently, KRL has become one of the most popular research areas and researchers have proposed many models to embed entities and relations in KGs. Next, we will introduce the typical models for KRL including linear model, neural model, translation model and other models.

\subsection{Linear Models}

Linear models employ a linear combination of the entities' and relations' representations to measure the probability of a fact triple. 

\subsubsection{Structured Embedding (SE)} SE \cite{bordes2011learning}  is one of the early models to embed KGs. 
 SE first learns relation-specific matrices  $\mathbf{M}_{r, 1}, \mathbf{M}_{r, 2} \in \mathbb{R}^{d\times d}$ for head entities and  tail entities respectively. After that, it multiples head and tail entities with the projecting matrix, and then defines the score function as $L_1$ distance between two multipled vectors for each triple $(h, r, t)$ as: 
\begin{equation}
f_r(h, t) =  \| \mathbf{M}_{r, 1} \mathbf{h} - \mathbf{M}_{r, 2} \mathbf{t} \|_{L_1}.
\end{equation}

 That is, SE transforms the entities' vectors $\mathbf{h}$ and $\mathbf{t}$ by the corresponding head and tail relation matrices for the relation $r$ and then measuring their similarities in the transformed relation specific space, which reflect the semantic relatedness of the head and tail entities in the relation $r$.

However, since the model learn two separate matrices for head and tail entities for each relation, it cannot precisely capture the semantic relatedness for entities and relations.
 
\subsubsection{Semantic Matching Energy (SME)} SME \cite{bordes2012joint,bordes2014semantic} 
first represents head entities, relations and tail entities with vectors respectively, and then models correlations between entities and relations  as semantic matching energy functions. SME defines a linear form for semantic matching energy functions:
\begin{equation}
f_r(h, t) = (\mathbf{M}_{1} \mathbf{h} + \mathbf{M}_{2} \mathbf{r} + \mathbf{b}_1 )^{\top} (\mathbf{M}_{3} \mathbf{t} + \mathbf{M}_{4} \mathbf{r} + \mathbf{b}_2),
\end{equation}
and also a bilinear form:
\begin{equation}
f_r(h, t) = \big( (\mathbf{M}_{1} \mathbf{h}) \otimes (\mathbf{M}_{2} \mathbf{r}) + \mathbf{b}_1 \big)^{\top} \big( (\mathbf{M}_{3} \mathbf{t}) \otimes (\mathbf{M}_{4} \mathbf{r}) + \mathbf{b}_2 \big),
\end{equation}
where $\mathbf{M}_{1}$, $\mathbf{M}_{2}$, $\mathbf{M}_{3}$ and $\mathbf{M}_{4} \in \mathbb{R}^{d\times d}$ are transformed matrices, $\otimes$ indicates the Hadamard product and $\mathbf{b}_1$ and $\mathbf{b}_2$ are bias vectors. In \cite{bordes2014semantic}, SME further extended it the bilinear form, which replace its matrices with  $3$-way tensors, to improve its model ability.

\subsubsection{Latent Factor Model (LFM)} LFM \cite{jenatton2012latent,sutskever2009modelling} employ a relation-specific bilinear form to consider the relatedness between entities and relations, and the score function for each triple $(h, r, t)$ is defined as:
\begin{equation}
f_r(h, t) = \mathbf{h}^{\top}\mathbf{M}_r\mathbf{t},
\end{equation}
where $M_r\in \mathbb{R}^{d\times d}$ are the matrix for relation $r$. 

It's a big improvement over the previous models since it interacts the distributed representations of head and tail entities  by a simple and efficient way. However, LFM is still restricting due to its massive number of parameters used for modeling the relations.

\subsubsection{DistMult}  DistMult \cite{yang2014embedding}  further reduce the number of relation parameters in  LFM, which simply restricts $\mathbf{M}_r$ to be a diagonal matrix. This results in a less complex model which achieves superior performance.

\subsubsection{ANALOGY} ANALOGY \cite{pmlr-v70-liu17d} uses the same bilinear form score function to measure the probability of fact triples as LFM and further discuss the normality and commutativity of LFM.

\subsection{Neural Models}

Neural Models aim to output the probability of the fact triples by neural networks which take the entities' and relations' embeddings as inputs.

\subsubsection{Multi Layer Perceptron (MLP)} MLP is proposed in \cite{dong2014knowledge} which employs a standard multi layer perceptron to capture interaction among entities and relations. The score function for each triple $(h, r, t)$ of MLP model is defined as:
\begin{equation}
f_r(h,t) = \mathbf{u}^\top\tanh(\mathbf{M}_1\mathbf{h}+\mathbf{M}_2\mathbf{r}+\mathbf{M}_3\mathbf{t})
\end{equation}
where $\mathbf{M}_1, \mathbf{M}_2, \mathbf{M}_3\in  \mathbb{R}^{d\times d}$ and $\mathbf{u}\in \mathbb{R}^d $ are the parameters of MLP.

\subsubsection{Single Layer Model (SLM)} SLM is similar to MLP model. It attempts to alleviate the issue of SE model by connecting entities and relations embeddings implicitly via the nonlinearity of a single MLP neural network. The score function for each triple $(h, r, t)$ of SLM model is defined as
\begin{equation}
f_{r}(h, t) = \mathbf{u}_r^\top \tanh (\mathbf{M}_{r, 1} \mathbf{h} + \mathbf{M}_{r, 2} \mathbf{t}),
\end{equation}
where $\mathbf{M}_{r, 1}, \mathbf{M}_{r, 2}\in \mathbb{R}^{k\times d}$ are weight matrices, $\mathbf{u}_r\in \mathbb{R}^{k}$ are the vector of relation $r$.

 Although SLM shows improvement over the SE model, it still suffer from problems when models large-scale KGs. The reason is that its non-linearity can only implicitly capture the interaction between entities and relations, and even lead to hard optimization. 

\subsubsection{Neural Tensor Network}
\begin{figure}[htb]
\centering
\includegraphics[width=0.6\columnwidth]{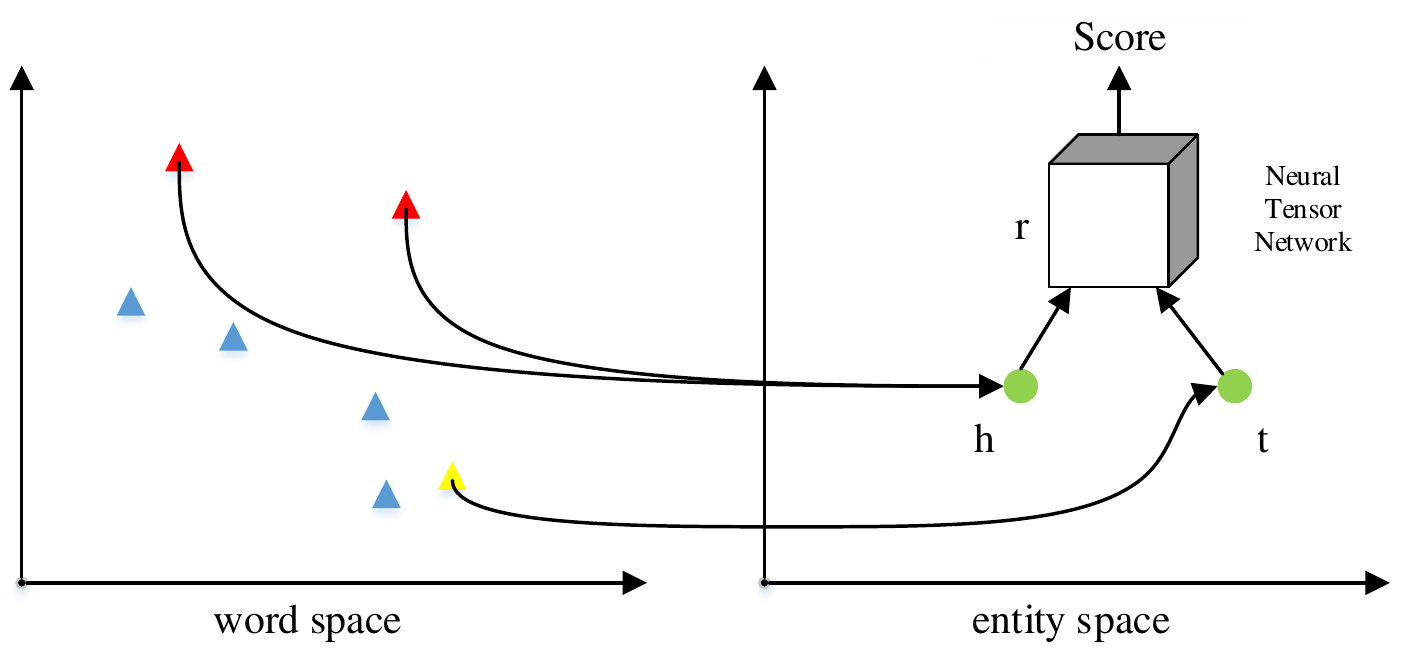}
\caption{Simple illustration of NTN.}
\label{fig:NTN}
\end{figure}
As illustrated in Figure \ref{fig:NTN},  Neural Tensor Network (NTN) \cite{socher2013reasoning} employs a bilinear tensor to combined two entities' embedding via multiple aspects.  The score function for each triple $(h, r, t)$ of NTN model is defined as:
\begin{equation}
f_{r}(h, t) = \mathbf{u}_r^\top \tanh (\mathbf{h}^{\top} \mathbf{M}_r \mathbf{t} + \mathbf{M}_{r, 1} \mathbf{h} + \mathbf{M}_{r, 2}\mathbf{t} + \mathbf{b}_r),
\end{equation}
where $\mathbf{M}_r \in \mathbb{R}^{d \times d \times k}$ is a $3$-way tensor, $\mathbf{M}_{r, 1}, \mathbf{M}_{r, 2} \in  \mathbb{R}^{k\times d}$ are weight matrices, and $\mathbf{u}_r$ is the vector of relation $r$. Note that,  SLM can be view as a special case of NTN without its tensor. 

Meanwhile, unlike previous KRL models modeling each entity with one vector, NTN represents each entity via averaging the word embeddings of their names. 
This approach can capture the semantic meaning of each entity name and further reduce the sparsity of entity representation learning. 

However, although the tensor operation in NTN can give a more explicit description of the comprehensive semantic relatedness between entities and relations, the following high complexity of NTN may restrict its applications on large-scale KGs.

\subsubsection{Neural Association Model (NAM)} NAM \cite{liu2016probabilistic} adopts multi-layer nonlinear activations in deep neural network to model the conditional probabilities between head and tail entities. NAM studies two model structures deep neural network (DNN) and relation modulated neural network (RMNN). 

NAM-DNN feeds the head and tail entities' embeddings into a MLP with $L$ fully connected layers, which is formalized as follows:
\begin{equation}
\mathbf{z}^l=\text{sigmoid}(\mathbf{M}^l\mathbf{z}^{l-1}+b^{l}),\ \ l = 1, \cdots, L,
\end{equation} 
where $\mathbf{z}^0=[\mathbf{h};\mathbf{r}]$, $\mathbf{M}^l$ and $\mathbf{b}^l$ is the weight matrix and bias vector for the $l$-th fully connected layer respectively. And finally the score function of NAM-DNN is defined as:
\begin{equation}
f_r(h,t) = \text{sigmoid}(\mathbf{t}^\top\mathbf{z}^L).
\end{equation}

Different from NAM-DNN, NAM-RMNN feds the relation embedding $\mathbf{r}$ into each layer of the deep neural network as follows:
\begin{equation}
\mathbf{z}^l=\text{sigmoid}(\mathbf{M}^l\mathbf{z}^{l-1}+\mathbf{B}^l\mathbf{r}),\ \ l = 1, \cdots, L,
\end{equation} 
where $\mathbf{z}^0=[\mathbf{h};\mathbf{r}]$, $\mathbf{M}^l$ and $\mathbf{B}^l$ indicate the weight matrices. And the score function of NAM-RMNN is defined as:
\begin{equation}
f_r(h,t) = \text{sigmoid}(\mathbf{t}^\top\mathbf{z}^L+\mathbf{B}^{l+1}\mathbf{r}).
\end{equation}

\subsection{Matrix Factorization Models}

Matrix factorization is an important technique to obtain low-rank representations. Hence, researchers also use matrix factorization in KRL.

A typical model of matrix factorization in KRL is {RESCAL}, a collective tensor factorization model presented in \cite{nickel2011three,nickel2012factorizing}, which reduce the modeling of the structure of KGs into a tensor factorization operation. In  {RESCAL}, the triples in KGs forms a large tensor $\mathbf{X}$ which $X_{hrt}$ is $1$ when $(h,r,t)$ holds, otherwise $0$. Tensor factorization aims to factorize $\mathbf{X}$ to entity embeddings and relation embeddings, so that $X_{hrt}$ is close to $\mathbf{h}\mathbf{M}_r\mathbf{t}$. Almost at the same time, \citep{drumond2012predicting} also use tensor factorization in KRL with the same way. We can find that {RESCAL} is similar to the previous model {LFM}. The major difference is that {RESCAL} will optimize all values in  $\mathbf{X}$ including the zero values while {LFM} focuses on the triples in KGs.

Besides RESCAL, there are other works utilizing matrix factorization in KRL. \citep{riedel2013relation,fan2014distant} learn representation for head-tail entity pair instead of single entity. Formally, it builds an entity-relation matrix $\mathbf{Y}$ which $Y_{ht,r}$ is $1$  when $(h,r,t)$ holds, otherwise $0$. And then matrix factorization is applied to factorize $\mathbf{Y}$ into entity pair embeddings and relation embeddings. Similarly, \citep{tresp2009materializing} and \citep{huang2014scalable} both model the head entity and the relation-tail entity pair with two separated vectors. However, such paired modeling cannot capture the interaction of the pairs and is easier to suffer from the issue of data sparsity.

\subsection{Translation Models}

Representation learning has been widely used in many NLP task since \citep{mikolov2013distributed} propose distributed word representation model and releases the tool word2vec. Mikolov et.al find some interesting phenomenon with their models. They find that the difference between the vectors of two words often embodies the relation between two words in the semantic space. For example, we have:
$$C(king) - C(queen)\approx C(man) - C(woman),$$
where $C(w)$ indicates the word vector of word $w$. In other words, word embeddings can capture the implicit semantic relatedness between $king$ and $queen$, $man$ and $woman$.
Moreover, they find that this phenomenon also exists in both lexical and syntactic relations according to their experimental results in the analogy task. Researchers\cite{fu2014learning} also use the features of word embeddings to discover the hierarchical relations between words.

\begin{figure}[htb]
\centering
\includegraphics[width=0.25\columnwidth]{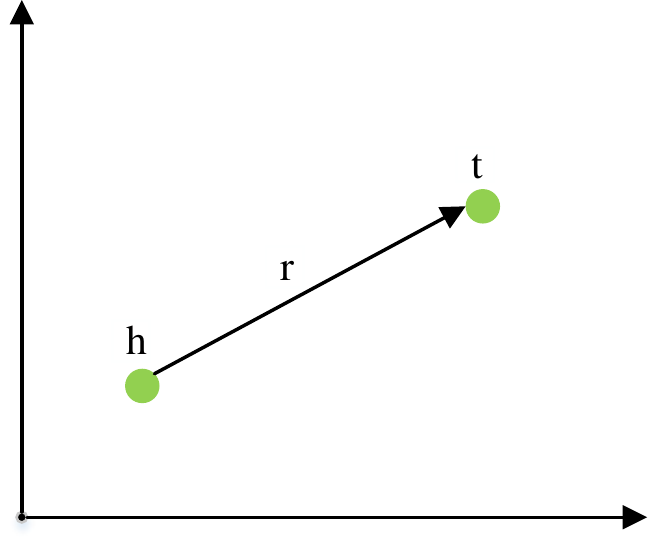}
\caption{A Simple illustration of TransE.}
\label{fig:TransE}
\end{figure}
Inspire by \cite{mikolov2013distributed}, \textbf{TransE} \cite{bordes2013translating} attempts to regard a relation $r$ as a translation vector $\mathbf{r}$ between the head and tail entities' vectors $\mathbf{h}$ and $\mathbf{t}$ for each triple $(h, r, t)$.  As illustrated in Figure \ref{fig:TransE}, TransE wants that $\mathbf{h} + \mathbf{r} \approx \mathbf{t}$ when $(h, r, t)$ holds. The score function for each triple $(h, r, t)$ is defined as:
\begin{equation}
\label{eq:transe}
f_r(h, t) =  \|\mathbf{h} + \mathbf{r} - \mathbf{t}\|_{L_1/L_2},
\end{equation}
where $f_r(h,t)$  can be either $L_1$ or $L_2$-norm.


Compared to traditional knowledge representation model, TransE can model complicated semantic relatedness between entities and relations with less model parameters and lower computational complexity. Bordes et al. evaluate the performance of TransE in the task of knowledge graph completion on the dataset of Wordnet and Freebase. The experimental results show that TransE has outperformed previous KRL models significantly, especially in the large-scale and sparse KGs.

Bordes et al.  also propose a naive version of TransE, the Unstructured Model \cite{bordes2012joint,bordes2014semantic}, which simply assigns zero vector for each relation, and the score function is defined as:
\begin{equation}
f_r(h, t) =  \|\mathbf{h} - \mathbf{t}\|_{L_1/L_2}.
\end{equation} 

However, due to the lack of relation embeddings, the Unstructured Model cannot consider relation information in the structure of KGs.

Since TransE is simple and efficient, many researchers expand TransE and apply it in many tasks. As it was, TransE is a typical model of knowledge representation. In the next section, we will take TransE as an example and introduce the major challenges and solutions in knowledge representation.

\subsection{Other Models}
Since the proposal of TransE, most of the new KR models have been based on it. Besides TransE and its extensions, we will also introduce some other models which also achieve promising performance.

\subsubsection{Holographic Embeddings (HolE)}
 To combine the expressive power of the tensor product with the efficiency and simplicity of TransE, HolE\cite{nickel2015holographic} uses the circular correlation of vectors to represent pairs of entities $a\star b $, where $\star$ : $\mathbb{R}_d \times \mathbb{R}_d \to \mathbb{R}_d$ denotes circular correlation:
\begin{equation}
[\mathbf{a}\star \mathbf{b}]_k = \sum_{i = 0}^{d-1} a_i b_{(i+k) mod d}.
\end{equation}

The circular correlation is not commutative and its single component can b viewed as a dot product operation. This makes it better model the 
 irreflexive relations and similar relations in KGs.  Moreover, although circular correlation can be interpreted as a special case of tensor product, it can be accelerated by fast Fourier transform  which makes it faster but maintains strong expressive ability. 



For each triple $(h, r, t)$, HolE define its score function as:
\begin{equation}
f_r(h,t) = \text{sigmoid}(\mathbf{r}^\top(\mathbf{h}\star\mathbf{t})).
\end{equation}

\subsubsection{Complex Embedding (ComplEx)} ComplEx \cite{welbl2016complex} employs eigenvalue decomposition model to take complex valued embeddings into consideration in KRL. The composition of complex embeddings makes ComplEx be capable of modeling  various kinds of binary relations. Formally, the score function of the fact $(h, r, t)$ of ComplEx is defined as:
\begin{equation}
f_r(h, t) = \text{sigmoid}(X_{hrt}),
\end{equation}
where $f_r(h, t)$ is expected to be $1$ when $(h, r, t)$ holds, otherwise $-1$.  Here, $X_{hrt}$ is further calculated as follows:
\begin{eqnarray}
X_{hrt} 
        &=& <\text{Re}(\mathbf{w}_{r}), \text{Re}(\mathbf{h}), \text{Re}(\mathbf{t})>\nonumber
       + <\text{Re}(\mathbf{w}_{r}), \text{Im}(\mathbf{h}), \text{Im}(\mathbf{t})>\nonumber\\
        & & -  <\text{Im}(\mathbf{w}_{r}), \text{Re}(\mathbf{h}), \text{Im}(\mathbf{t})>\nonumber
              -  <\text{Im}(\mathbf{w}_{r}), \text{Im}(\mathbf{h}), \text{Re}(\mathbf{t})>,
\end{eqnarray}
where $\mathbf{M}_r\in \mathbb{R}^{d\times d}$ is a weight matrix, $<\mathbf{a},\mathbf{b},\mathbf{c}> = \sum_k a_kb_kc_k$, $Im(x)$ indicates the the imaginary part of $x$ and $Re(x)$ indicates  the the real part of $x$ . Note that, ComplEx can be view as an extension of RESCAL, which assigns  complex embedding of the entities and relations.

Besides, \citep{hayashi2017equivalence} have proved that HolE is mathematically equivalent to ComplEx recently.

\section{The Main Challenges of Knowledge Graph Representation Learning}

Recently, knowledge representation models such as TransE have achieved significant improvement in many real-world tasks.  However, there are still many challenges in the KRL. In this section, we will take TransE as an example and introduce some related works which try to solve the problems in KRL.

\subsection{Complex Relation Modeling}
TransE is simple and effective, which has promising performance in large-scale KGs. However, due to TransE's simpleness,  it cannot deal with the modeling of complex relations in KGs.

Here, complex relations are defined as follows. According to their mapping properties, the relations are divided into four types including 1-to-1, 1-to-n, n- to-1 and n-to-n relations. Take 1-to-n relation as example, it means that the head entity in this relation links with multiple tail entities. We regard 1-to-n, n-to-1 and n-to-n relations as complex relations.

Researchers have found that existing KRL models have poor performance when dealing with complex relations. Take TransE as an example, since TransE regards a relation $r$ as a translation vector between head and tail entity. 

it hopes $\mathbf{h} + \mathbf{r} \approx \mathbf{t}$ for each fact triple $(h, r, t)$. Therefore, we will obtain the following contradiction directly:

(1) If the relation r is a reflexive relation such as \texttt{friends}, i.e.,  $(h, r, t)\in S$ and  $(t, r, h)\in S$, we will get $\mathbf{r}\approx \mathbf{0}$ and $\mathbf{h}\approx\mathbf{t}$.

(2) If the relation r is a 1-to-n relation, i.e. $\forall i\in\{0,1,\cdots,m\}, (h, r ,t_i)\in S$, we will get $\mathbf{t}_0\approx\mathbf{t}_1\approx\cdots\approx\mathbf{t}_m$. Similarly, this problem also exists for the situation when $r$ is a n-to-1 relation.

\begin{figure}[h!]
\centering
\includegraphics[width=0.4\columnwidth]{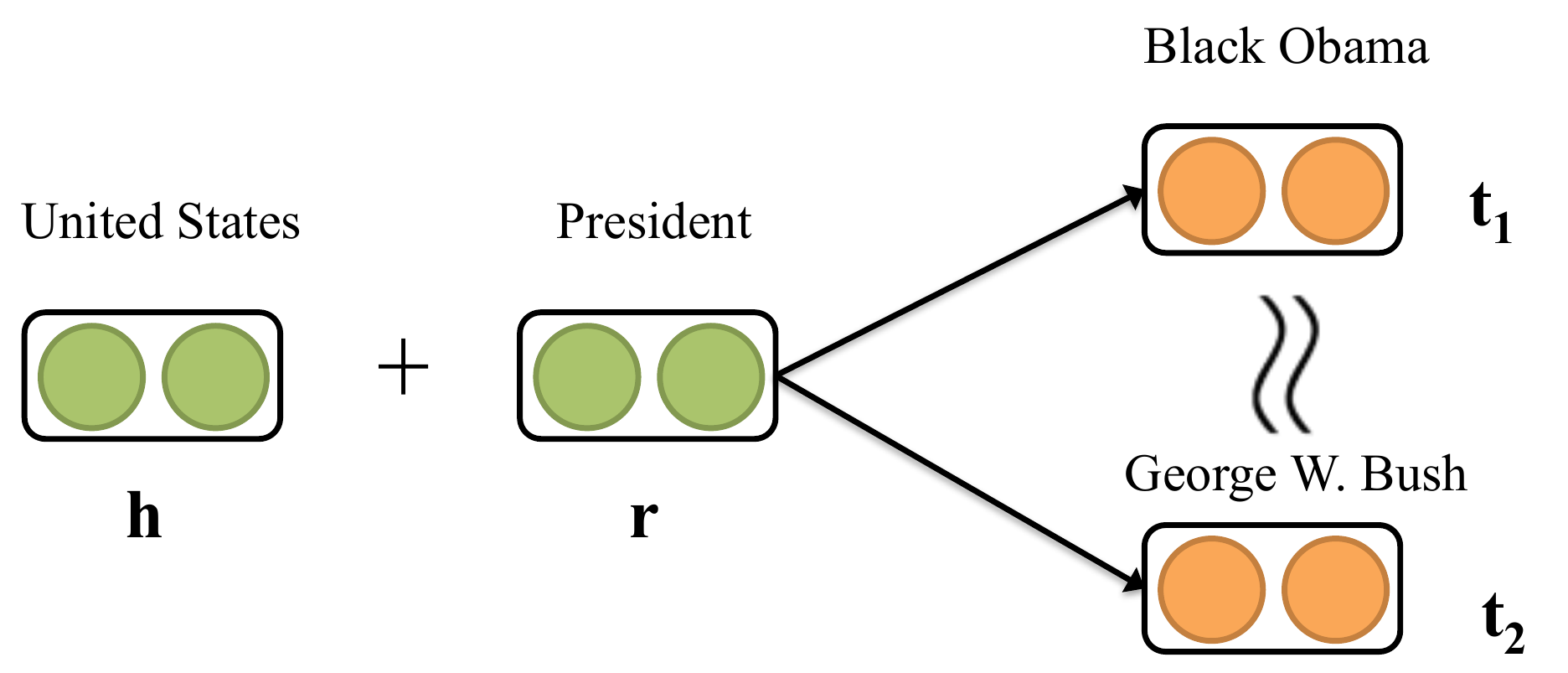}
\caption{The example of complex relations.}
\label{fig:complex-relations}
\end{figure}

For example, there are two triples (\emph{United States}, \texttt{President}, \emph{Black Obama}) and  (\emph{United States}, \texttt{President}, \emph{George W. Bush}) in KGs.  Here, the relation \texttt{President} is a typical one-to-many relation. If we use  TransE to model these two triples, as illustrated in Figure \ref{fig:complex-relations}, we will get the same embeddings of  \emph{Black Obama} and \emph{George W. Bush}.

This obviously deviates from the truth. \emph{Black Obama} and \emph{George W. Bush} varies in many aspects except that they are both presidents of \emph{United States}. Therefore,  the entity embeddings gained by TransE are lacking in discrimination due to these complex relations.

Hence, how to deal with complex relations is one of the main challenges in KRL. Recently, there are some extensions of TransE which focus on this challenge. We will introduce these models in this section.

\subsubsection{TransH}
To address the issue of TransE when modeling complex relations, TransH \cite{wang2014knowledge} is proposed that an entity should have different distributed representations in the fact triples with different relations.

\begin{figure}[htb]
\centering
\includegraphics[width=0.3\columnwidth]{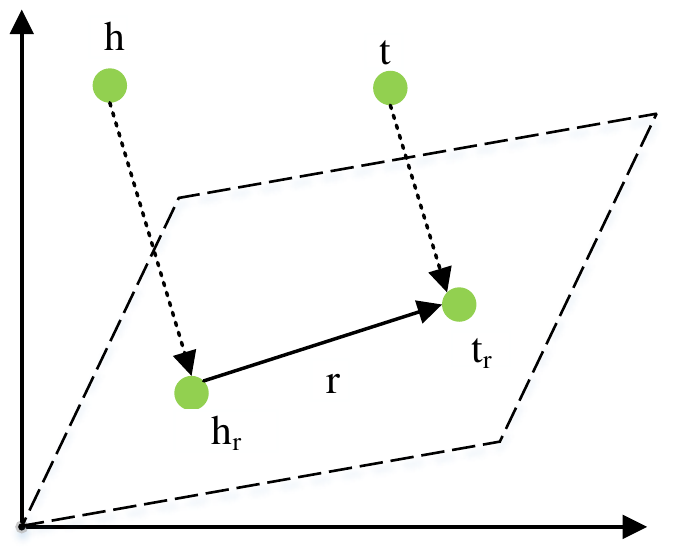}
\caption{Simple illustration of TransH.}
\label{fig:TransH}
\end{figure}

As illustrated in Figure \ref{fig:TransH}, for a relation, TransH projects head and tail entities into the specific hyperplane of this relation. Formally, 
for a triple $(h, r, t)$, the head and tail entity are first projected to the hyperplane of the relation $r$, denoted as $\mathbf{h}_{r}$ and $\mathbf{t}_{r}$ which is calculated by:
\begin{equation}
\mathbf{h}_{r} = \mathbf{h} - \mathbf{w}_{r}^{\top}\mathbf{h}\mathbf{w}_{r},\ \ \
 \mathbf{t}_{r} = \mathbf{t} - \mathbf{w}_{r}^{\top}\mathbf{t}\mathbf{w}_{r},
\end{equation}
where $\mathbf{w}_r$ is the normal vector of the hyperplane.  Then the score function for each triple $(h, r, t)$ is defined as
\begin{equation}
f_{r}(h, t) = \|\mathbf{h}_{r} + \mathbf{r} - \mathbf{t}_{r}\|_{L_1/L_2}.
\end{equation}

Note that, there may exist infinite number of hyperplanes for a relation $r$, but TransH simply requires $\mathbf{r}$ and $\mathbf{w_r}$ to be approximately orthographic by restricting $\|\mathbf{w}_r\mathbf{r}\|_{L_2} = 0$,

\subsubsection{TransR/CTransR}
Although TransH enables an entity having different representations for different relations, it still simply assumes that entities and relations can be represented in a unified semantic space.  It prevents TransH from modeling entities and relations precisely. TransR \cite{lin2015learning} observes that an entity may exhibit its different attributes in distinct relations and models entities and relations in separated spaces. As a result, although some entities such as \emph{Beijing} and \emph{London}  are far away from each other in entity space, they are similar and close to each other  in the some specific relation spaces, and vice versa.
\begin{figure}[htb]
\centering
\includegraphics[width=0.6\columnwidth]{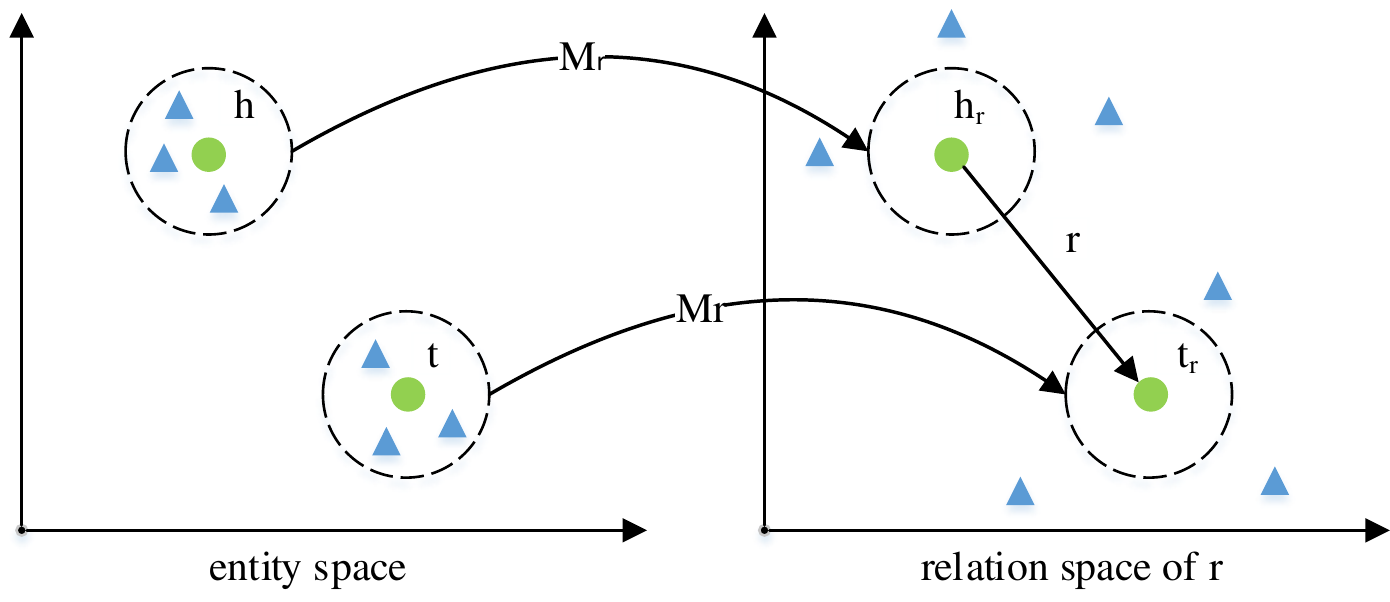}
\caption{Simple illustration of TransR. }
\label{fig:TransR}
\end{figure}

As illustrated in Figure \ref{fig:TransR}, for each triple $(h, r, t)$,  non-relevant head/tail entities (denoted as colored triangles) are kept away from relevant entities (denoted as colored circles) in the specific relation space by relation-specific projection, meanwhile these entities are not necessarily far away from each other in entity space.

For each triple $(h, r, t)$, TransR first projects head and tail entities  from entity space to $r$-relation space via a projection matrix $\mathbf{M}_{r} \in \mathbb{R}^{d \times k}$, denoted as $\mathbf{h}_r$ and $\mathbf{t}_r$, which is defined as:
\begin{equation}
\mathbf{h}_{r} = \mathbf{h}\mathbf{M}_r, \ \ \ \mathbf{t}_{r} = \mathbf{t}\mathbf{M}_r.
\end{equation}

And then we force that $\mathbf{h}_r + \mathbf{r} \approx \mathbf{t}_r$. For each triple $(h, r, t)$, the score function is correspondingly defined as:
\begin{equation}
f_{r}(h, t) = \|\mathbf{h}_r + \mathbf{r} - \mathbf{t}_r\|_{L_1/L_2}.
\end{equation}

Besides, \citep{nguyen-EtAl:2016:N16-1} propose an extension of TransR: STransE that represents a relation with two different mapping matrices and a translation vector.

Further, Lin et al. found that a specific relation usually corresponds to head-tail entity pairs with distinct attributes. For example, for the relation ``/location/location/contains'', its head-tail entities pattern may be continent-country, country-city, country-university,  and so on. If current relations are divided into more precise sub-relations, the entities can be projected into a more accurate sub-relation space. It should be beneficial to represent KGs.

Therefore, Lin et al.  propose CTransR which clusters all triples $(h, r,  t)$ involved for a specific relation $r$  into multiple groups according to the embedding offsets $\mathbf{h}-\mathbf{t}$. And the relations in the triples of the same group are defined as a new sub-relation. Then CTransR learns a sub-relation vector $\mathbf{r}_c$  and relation-specific projection matrix $M_{r, c}$ for each cluster. For each triple $(h, r, t)$, the score function of CTransR is finally defined as
\begin{equation}
f_{r}(h, t) = \|\mathbf{h}_{r,c} + \mathbf{r_c} - \mathbf{t}_{r,c}\|_{L_1/L_2},
\end{equation}
where $\mathbf{h}_{r,c} = \mathbf{h}\mathbf{M}_{r, c}$ and $\mathbf{t}_{r,c} = \mathbf{t}\mathbf{M}_{r, c}$.

\subsubsection{TransD}
In fact, although TransR has significant improvements compared with TransE and TransH, it still has several limitation. First, it simply share relation-specific projection matrix in head and tail entities, ignoring various types and attributes of head and tail entities. Moreover, as compared to TransE and TransH, TransR has much more parameter and higher computation complexity due to its matrix multiplication operation.



To address these issues, \citep{jiknowledge2015} propose TransD which sets different mapping matrices for head and tail entities.
\begin{figure}[htb]
\centering
\includegraphics[width=0.65\columnwidth]{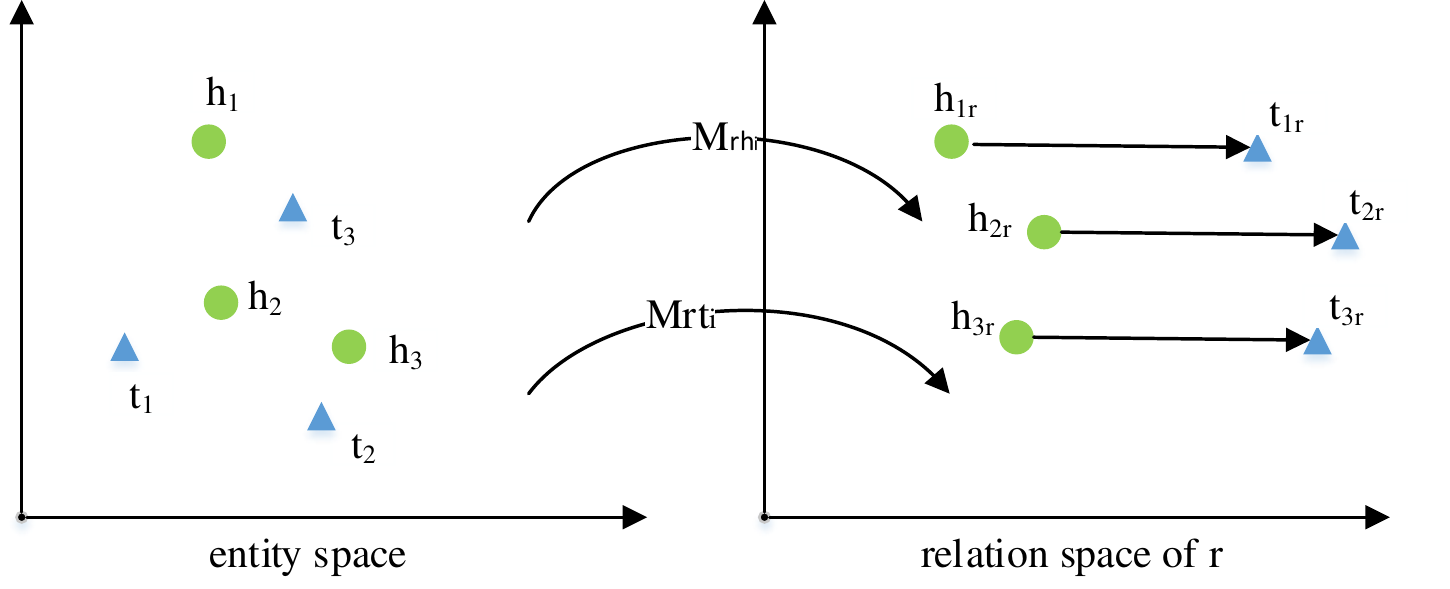}
\caption{Simple illustration of TransD. $\mathbf{M}_{rh_i}$ and $\mathbf{M}_{rt_i}$ are mapping matrices of $h_i$ and $t_i$, respectively}
\label{fig:TransD}
\end{figure}

As illustrated in Figure \ref{fig:TransD}, for a triple $(h, r, t)$, TransD further learns two projecting matrices $M_{rh}$, $M_{rt} \in R^{d\times k}$ to project head and tail entities from entity space to relation space respectively, which are defined as follows:
\begin{equation}
\mathbf{M}_{rh} = \mathbf{r}_p\mathbf{h}^\top_p + \mathbf{I}^{d\times k}, \ \ \ 
\mathbf{M}_{rt} = \mathbf{r}_p\mathbf{t}^\top_p + \mathbf{I}^{d\times k},
\end{equation}
where $\mathbf{I}\in\mathbb{R}^{d\times d}$ indicates the identical matrix, $\mathbf{h}$, $\mathbf{h}_p$  , $\mathbf{t}$, $\mathbf{t}_p\in \mathbb{R}^d$ , $\mathbf{r}$, $\mathbf{r_p}\in \mathbb{R}^k$ and subscript $p$ marks the projection vectors. Here, the mapping matrices $\mathbf{M}_{rh}$ and $\mathbf{M}_{rt}$ are related to both entities and relations, and using two projection vectors instead of matrices solves the issue of large amount of parameter in TransR. Hence, for a triple $(h, r, t)$, the score function of TransD is defined as:
\begin{equation}
f_r(h,t) = \|\mathbf{hM}_{rh} + \mathbf{r} - \mathbf{tM}_{rt}\|_{L_1/L_2}.
\end{equation}

Further, \citep{yoon2016translation}  propose a  KRL model based on TransE, TransR, and TransD to  preserve the logical properties of relations.

\subsubsection{TranSparse}
Although existing translation-based models have strong ability to model KGs, they are still far from practicality since entities and relations are heterogeneous and unbalanced, which is a great challenge in KRL. 



To address these two issues, TranSparse \cite{ji2016knowledge}  considers the heterogeneity and the imbalance when modeling entities and relations in KGs. To overcome the heterogeneity, TranSparse(share) which replaces the dense matrices  in TransR with sparse matrices, of which the sparse degrees is determined by he number of entity pairs related to corresponding relations. Formally, for each relation $r$, the projection matrix $\mathbf{M}_r(\theta_r)$'s sparse degree is $\theta_r$ which is defined as:
\begin{equation}
\theta_r = 1 - (1 - \theta_{min})N_r/N_{r^*},
\end{equation}
where $0 \leq \theta_{min}\leq 1$ is a hyper-parameter indicating the minimun sparse degree, $N_r$ indicates the number of entity pairs related to relation $r$, and $r^*$ is the relation which relates to the most entity pairs. Therefore, the projected entity vectors can be calculated by:
\begin{equation}
\mathbf{h}_r = \mathbf{M}_r(\theta_r)\mathbf{h},\ \ \ 
\mathbf{h}_t = \mathbf{M}_r(\theta_r)\mathbf{t}.
\end{equation}

Besides, TranSparse(seperate) uses two different projection matrices $\mathbf{M}_r^h(\theta_r^h)$ and $\mathbf{M}_r^t(\theta_r^t)$ for head entity $h$ and tail entity $t$ to deal with the issue of imbalance of relations. The sparse degree is defined as:
\begin{equation}
\theta_r^l = 1 - (1 - \theta_{min})N_r^l/N_{r^*}^{l^*},
\end{equation}
where $N_r^l$ denotes the number of head/tail entities related to relation, and $N_{r^*}^{l^*}$ denotes the maximum one in $N_r^l$.

Hence, the projection vector of head/tail entities is defined as:
\begin{equation}
\mathbf(h_r) = \mathbf{M}_r^h(\theta_r^h)\mathbf{h}, \ \ \ 
\mathbf(h_t) = \mathbf{M}_r^t(\theta_r^t)\mathbf{t}.
\end{equation}

And for both TranSparse(share) and  TranSparse(seperate), the score function  for a triple $(h, r, t)$ of TranSparse is defined as:
\begin{equation}
f_{r}(h, t) = \|\mathbf{h}_r + \mathbf{r} - \mathbf{t}_r\|_{L_1/L_2}.
\end{equation}

\subsubsection{TransA}
\citep{xiao2015transa} think that TransE and its extensions have two major problems: (1) TransE and its extensions only use $L_1/L_2$ distance in their loss metric. Hence, they are lacking in flexibility. (2) TransE and its extensions treat each dimension of entities and relations vectors identically due to the oversimplified loss metric.

To address these two issues, TransA is proposed to change the oversimplified loss metric and to replace inflexible $L_1$ or $L_2$ distance with adaptive Mahalanobis distance of absolute loss. The score function of TransA is defined as follows:
\begin{equation}
f_r(h,t) = (\mathbf{h} + \mathbf{r} - \mathbf{t})\mathbf{M_r}(\mathbf{h} + \mathbf{r} - \mathbf{t})^\top
\end{equation}
where  $\mathbf{M_r}$ is a relation-specific symmetric non-negative weight matrix that corresponds to the adaptive metric.
\begin{figure}[htb]
\centering
\includegraphics[width=0.65\columnwidth]{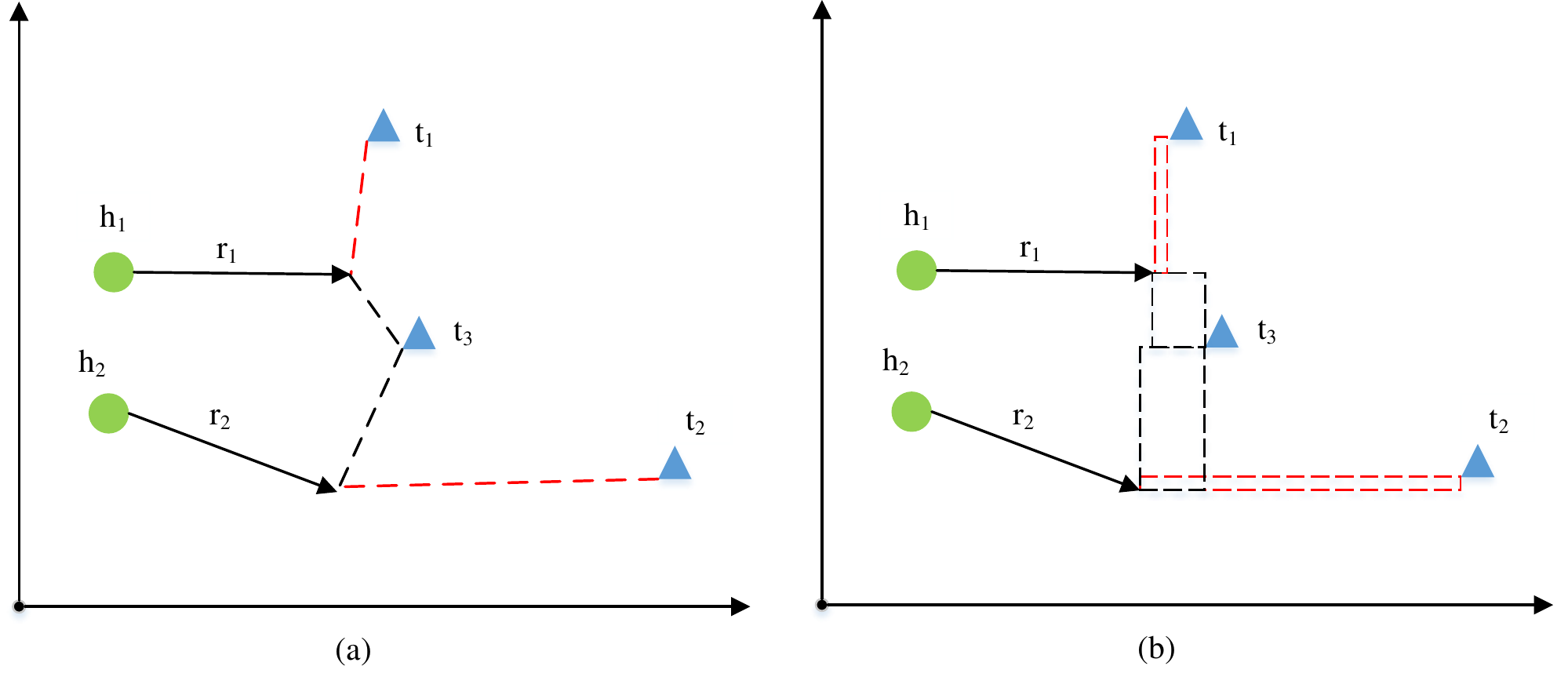}
\caption{Simple illustration of TransA. }
\label{fig:TransA}
\end{figure}

As illustrated in Fig. \ref{fig:TransA}, $t_1$ and $t_2$ are correct tail entities while $t_3$ are not. Fig. \ref{fig:TransA}(a) shows that the incorrect entity is matched with the L2-norm distance. And Fig. \ref{fig:TransA}(b) shows that by weighting embedding dimensions, the embeddings are refined because the correct entities have a smaller loss in x-axis or y-axis direction.

Similar to TransA, TransM \cite{fan2014transition} also proposes a new loss function in KRL, which assigns each fact triple $(h, r, t)$ with a relation-specific weight $\theta_r$. The key idea of TransM is that different relation may have different importances when learning the representations of KGs. And the score function of TransM is defined as:
\begin{equation}
f_r(h,t) = \theta_r||\mathbf{h}+\mathbf{r}-\mathbf{t}||_{L_1/L_2}.
\end{equation}
TransM alleviates incapability of modeling complex relations in TransE by assigning lower weights to those relations.

Besides, TransF \cite{feng2016knowledge} employs dot product instead of the $L_1$ or $L_2$ distance in TransE to measure the probability of fact triple $(h, r, t)$, and the score function of TransF is defined as:
\begin{equation}
f_r(h,t) = (\mathbf{h}+\mathbf{r})^\top\mathbf{t}+(\mathbf{t}-\mathbf{r})^\top\mathbf{h}.
\end{equation}

That is, TransF wants the vector of head entity $\mathbf{h}$ to have the same direction with $\mathbf{h}+\mathbf{r}$, and the vector of tail entity $\mathbf{h}$ to have the same direction with $\mathbf{t}-\mathbf{r}$. 

\subsubsection{TransG} 
\begin{figure}[htb]
\centering
\includegraphics[width=0.65\columnwidth]{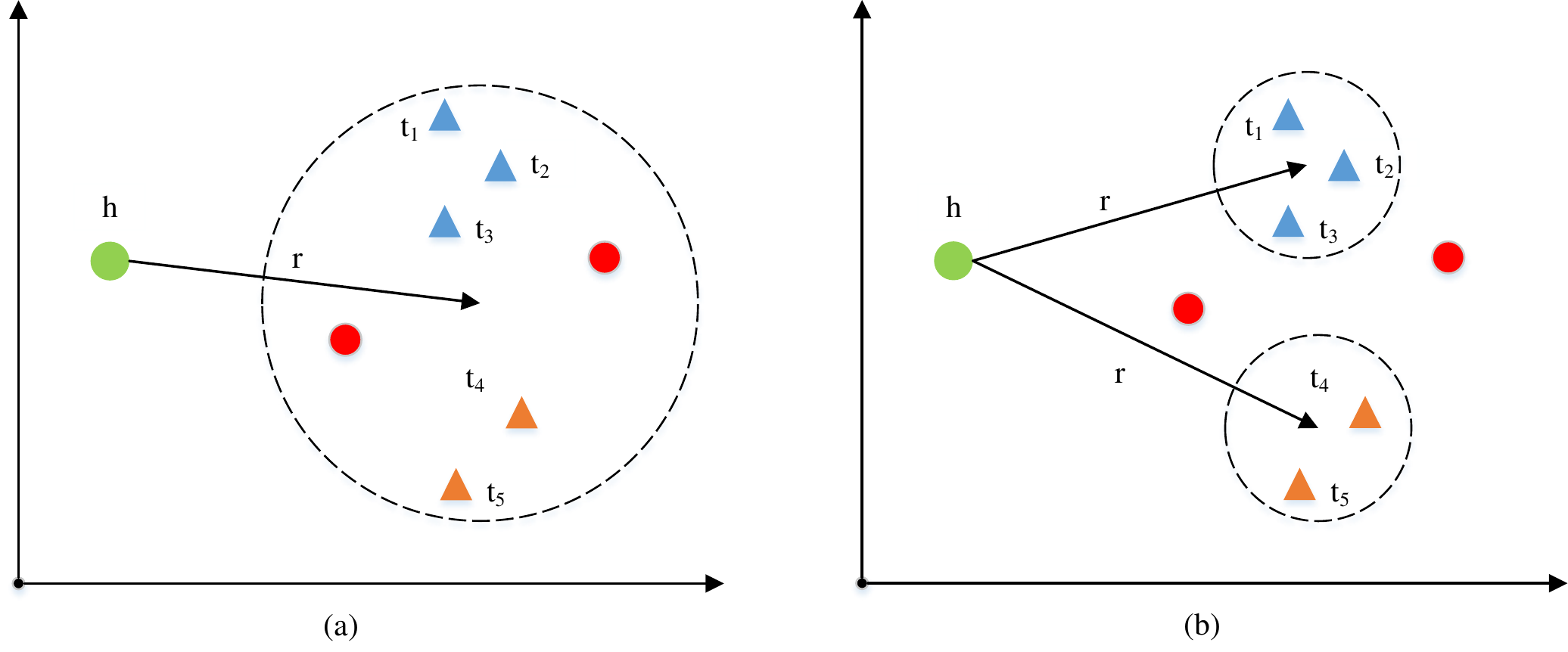}
\caption{Simple illustration of TransG. The triangles indicates the valid tail entities for  $(h, r)$ , while the circles indicates the invalid ones. }
\label{fig:TransG}
\end{figure}

Similar to CTransR, TransG \cite{xiao2015transg} finds that existing translation-based models such as TransE cannot deal with the situation that  a relation has multiple meanings when involves with different entity pairs. The reason is that these models only maintain a single vector for each relation, which may be insufficient to model distinct relation meanings. As illustrated in Fig. \ref{fig:TransG}(a) shows that the valid triples cannot be distinguished from the incorrect ones by   existing translation-based models since all semantic meanings of relation $r$ are regarded as the same.  Fig. \ref{fig:TransG}(b) shows that by considering the multiple semantic meanings of relations, TransG model could discriminate the valid triples from the invalid ones. 

TransG proposes to use Bayesian non-parametric infinite mixture embedding to take the multiple semantic meanings of relations into consideration in KRL. For each entity, TransG assumes that the entity embedding vector subjects to standard normal distribution, i.e., 
\begin{equation}
\mathbf{h}\sim \mathcal{N}(\mathbf{u}_h, \sigma_h^2\mathbf{I}),\ \ \ 
\mathbf{t} \sim \mathcal{N}(\mathbf{u}_t, \sigma_t^2\mathbf{I}),
\end{equation}
where $\mathbf{I}\in \mathbb{R}^{d\times d}$  indicates the identical matrix, $\mathbf{u}_h, \mathbf{u}_t\sim\mathcal{N}(\mathbf{0}, \mathbf{1})$ is the mean of head and tail entity vectors respectively, $\sigma_h, \sigma_t$ indicate the variance of head and tail entity vectors' distribution respectively. Hence,  the relation vector is then defined as 
\begin{equation}
\mathbf{r}_i=\mathbf{t}-\mathbf{h}\sim\mathcal(\mathbf{u}_t-\mathbf{u}_h, (\sigma_h^2+\sigma_t^2)\mathbf{I}).
\end{equation}
where $\mathbf{r}_i$ indicates  the relation  embedding  vector  for the $i$-th  semantic meaning of relation $r$.

Then for a triple $(h, r, t)$, the score function of TransG is defined as:
\begin{eqnarray}
f_r(h,t) &=& \sum_{i=1}^{M_r}\pi_{r,i} e^{\frac{-||\mathbf{h}+\mathbf{r}_i-\mathbf{t}||^2_2}{\sigma_h^2+\sigma_t^2}}.
\end{eqnarray}
where $\pi_{r,i}$ is   the  weight  factor corresponding to $i$-th semantic meaning of relation $r$.


\subsubsection{KG2E} He at el. \cite{he2015learning}  notice that the semantic meanings of entities and relations in KGs are often uncertain. However, previous translation-based models do not consider this phenomenon when distinguishing a valid triple and its corresponding invalid triples. In order to explicitly consider KG's uncertainties, KG2E represent entities or relations in KG through a vector with Gaussian distribution instead of a single vector. For an entity or a relation, they want the mean of its embedding to denote the center position of its semantic meanings, and the covariance matrix to describes its uncertainty.

\begin{figure}[htb]
\centering
\includegraphics[width=0.45\columnwidth]{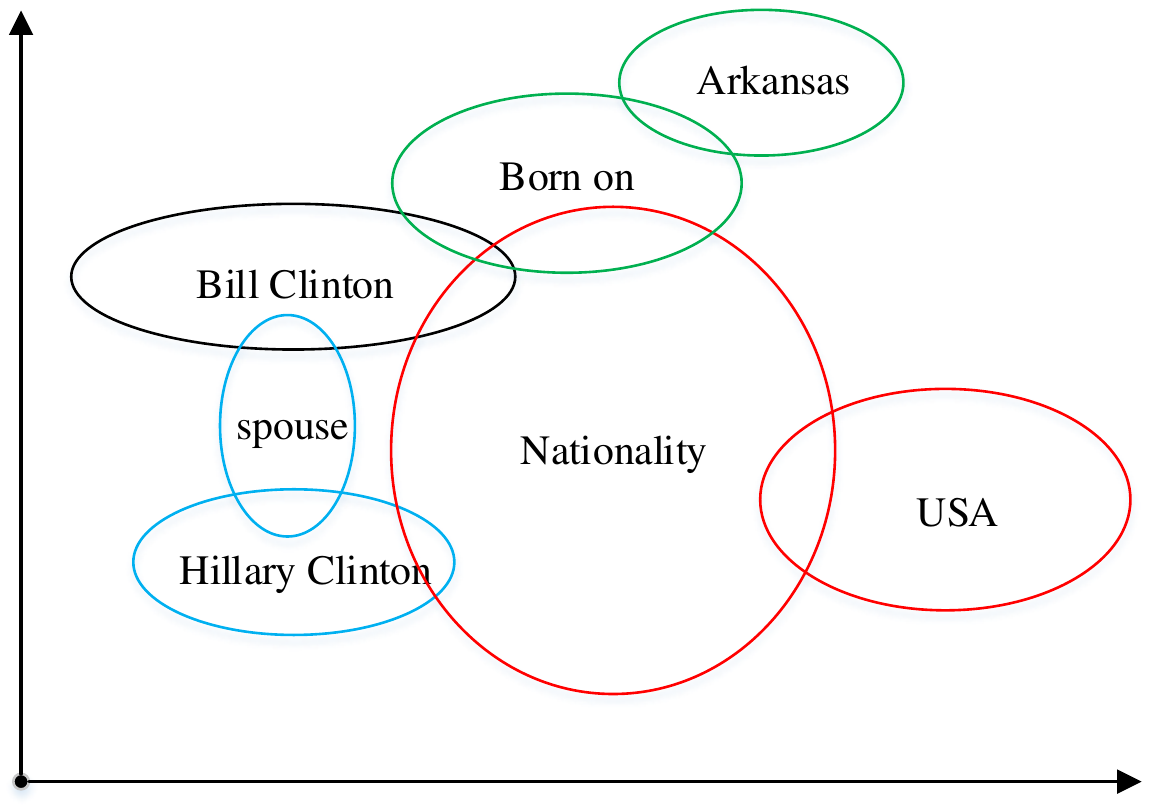}
\caption{Simple illustration of KG2E. KG2E represents entities and relation with Gaussian Embedding.}
\label{fig:KG2E}
\end{figure}

As illustrated in Figure \ref{fig:KG2E}, each circle denotes an entity or a relation in KG, while its size denotes the corresponding uncertainty. Here, we can find that the uncertainty of relation \texttt{Nationality}  is higher than other relations.

KG2E uses $\mathbf{h} - \mathbf{t}$ to express the relation between head entity $h$ and tail entity $t$, which corresponds to the probability distribution $P_e$:
\begin{equation}
P_e \sim N (\mu_h - \mu_t , \Sigma_h + \Sigma_t ).
\end{equation}

And the relation $r$ can be also expressed by a probability distribution of relation $P_r\sim N (\mu_r , \Sigma_r )$. Hence, we can measure the similarity of triple $(h, r, t)$ by measuring the similarity of two distribution $P_e$ and $P_r$.  In KG2E, the similarity between $P_e$ and $P_r$ is defined with two measures: KL-divergence and expected likelihood.

(1) Asymmetric similarity: KL-divergence based score function (KG2E\_KL) is defined as
\begin{eqnarray}
f_r(h,t) &=& \int_{x\in R^{k_e}} N(x;\mu_r, \Sigma_r) \log \frac{N(x;\mu_e, \Sigma_e)}{N(x;\mu_r, \Sigma_r)} dx \nonumber\\
&=& \frac{1}{2}\big\{tr(\Sigma_r^{-1}\Sigma_e)+(\mu_r-\mu_e)^\top\Sigma_r^{-1}(\mu_r-\mu_e)-\log\frac{\det(\Sigma_e)}{\det(\Sigma_r)}-k_e\big\}.
\end{eqnarray}

(2) Symmetric similarity: expected likelihood based score function (KG2E\_EL) is defined as
\begin{eqnarray}
f_r(h,t) &=& \int_{x\in R^{k_e}} N(x;\mu_r, \Sigma_r) N(x;\mu_e, \Sigma_e) dx \nonumber\\
&=& \frac{1}{2}\big\{(\mu_e-\mu_r)^\top(\Sigma_e+\Sigma_r)^{-1}(\mu_e-\mu_r)+\log \det(\Sigma_e + \Sigma_r)+k_e\log(2\pi)\big\}.
\end{eqnarray}

Note that, to avoid overfitting, KG2E needs regularization during learning. It uses the following hard constraints:
\begin{equation}
\forall l\in E\bigcup R, c_{min}\mathbf{I}\leq \Sigma_l\leq c_{max}\mathbf{I}, c_{min}>0
\end{equation}

\subsubsection{ManifoldE} \citep{xiao2016knowledge} discover that existing KRL models could not make a precise knowledge graph completion in large-scale KG because of these models all employ  an overstrict geometric form and  an ill-posed algebraic system.   To address these issue, they propose a novel model ManifoldE, which adopts manifold-based embedding principle instead of traditional $L_1$ or $L_2$ distance to model fact triples. Hence, for a given fact triple, the score function of ManifoldE is calculated by measuring the distance in the manifold:
\begin{equation}
f_r(h, t) = ||\mathcal{M}(h,r,t)-D_r^2||^2,
\end{equation}
where $D_r$ is a relation-specific parameter and $\mathcal{M}$ is the manifold function which can be defined in two different ways:

\textbf{Sphere} assumes that for a fact triples, its head and tail entities lay in a sphere with radius $D_r$. Hence, $\mathcal{M}$ is defined as:
\begin{equation}
\mathcal{M}(h,r,t) = ||\mathbf{h}+\mathbf{r}-\mathbf{t}||_2^2.
\end{equation}

\textbf{Hyperplane} proposes to embed head and tail entities into two separated hyperplanes, and intersect with each other when their hyperplanes are not parallel. Hence, $\mathcal{M}$ is defined as:
\begin{equation}
\mathcal{M}(h,r,t) = (\mathbf{h}+\mathbf{r}_h)^\top(\mathbf{t}+\mathbf{r}_t),
\end{equation}
where $\mathbf{r}_h$ and $\mathbf{r}_t$ are specific relation vectors of head and tail entities respectively.

Recently, TransE's extensions such as TransH, TransR, TransD, TranSparse, TransA, TransG, KG2E and so on have invested in dealing with the complex relation modeling issue. The experimental results on real-world datasets show that these methods have improvements as compared to TransE, which reveals the effectiveness of these models to consider different characteristics of the complex relations in KGs.

\subsection{Relational Path Modeling}

Although TransE and its extensions has achieved the great success in modeling entities and relations in KGs,  they still face a problem caused by only considering direct relations between entities. It is known that, there are also relational paths between entities, which indicates the complicated semantic relatedness between entities. For example, the relation path $h \xrightarrow{\texttt{Father}} e_1 \xrightarrow{\texttt{Mother}} t$ indicates there is a relation \texttt{GrandMother} between $h$ and $t$, i.e., ($h$, \texttt{GrandMother}, $t$). In fact, relational paths have been taking into consideration in knowledge inference on large-scale KGs. \citep{lao2010relational,lao2012reading,lao2011random,gardner2013improving} propose Path Ranking Algorithm (PRA) and apply it in finding unknown relational facts in large-scale KGs. PRA uses the relational paths between entities to predict their relations, and achieves great success, which indicates that relational paths between entities are informative for infer unknown facts.

Inspired by PRA algorithm, \citep{lin2015modeling} propose Path-based TransE (PTransE) which extends TransE to model relational paths in KGs. Since the large number of relational paths in KGs and they usually contain noises, PTransE utilizes a  Path-Constraint Resource Allocation (PCRA) algorithm to measure if a relational path is reliable. Further, PTransE proposes three typical operation including addition, multiplication and Recurrent neural network (RNN) to compose the relation embeddings into relational path embedding. Formally,  for a relational path $p = (r_1, \ldots, r_l)$, the addition operation which is formalized as:


$\mathbf{p} = \mathbf{r}_1 + \ldots + \mathbf{r}_l,$
and the multiplication which is formalized as: $
\mathbf{p} = \mathbf{r}_1 \cdot \ldots \cdot \mathbf{r}_l,$

and  the composition operation of RNN is defined using a reccurent matrix :$
\mathbf{c}_i = f(\mathbf{W}[\mathbf{c}_{i-1}; \mathbf{r}_i])$, and the relational path embedding is defined as the final state pf RNN $\mathbf{p} = \mathbf{c}_n.$

Finally, the score function of PTransE is defined as:
\begin{equation}
f_r(h,t) = \frac{1}{Z} \sum_{p \in P(h, t)} R(p) ||\mathbf{p} - \mathbf{r}||_{L_1/L_2} + ||\mathbf{h}+\mathbf{r}-\mathbf{t}||_{L_1/L_2},
\end{equation}
where $P(h,t)$ indicates the set of relational paths found by PCRA,  $R(p)$ indicates the reliability of the relational path $p$ and $Z$ is a normalized constant.

Almost at the same time, there are other researchers considering relational paths with a similar way in KRL successfully \cite{garciacomposing,neelakantan-roth-mccallum:2015:ACL-IJCNLP,luo2015context}. Algorithms utilizing the information of relational paths always suffer from expensive computation cost induced by enumerating paths between entities. Both \cite{lin2015modeling} and \cite{garciacomposing} address this issue by sampling informative paths. \cite{toutanova-EtAl:2016:P16-1} propose to utilize dynamic  programming algorithm to make use of all relation paths efficiently. And \cite{das-EtAl:2017:EACLlong1} propose to use attention mechanism to incorporate multiple relational paths. Further, \citep{feng-EtAl:2016:COLING1} propose to leverage the graph’s structure information into KRL. Besides, relational path learning has also been used in relation extraction \cite{zeng-EtAl:2017:EMNLP2017} and KG-based QA \cite{gu2015traversing}.

The successes in PTransE and other related models have shown that taking relational paths into accounts can significantly improve the discrimination of relational learning and the system performance in the task of knowledge graph completion and so on. However, the existing models are still some preliminary attempts at modeling relational paths. There are many further investigations in the reliability measure and semantic composition of relational paths to be done.

\subsection{Multi-source Information Learning}

Most KRL methods stated above only concentrate on the fact triples themselves in the KG, regardless of the rich multi-source information such as textual information, type information, visual information and so on. This cross-modal information could provide additional knowledge located in plain texts, type structures or figures of entities and is important when learning knowledge graph representations.

\subsubsection{Textual Information} Textual information is one of the most significant and widely spread information we send out and receive in every day. It is intuitive that we can consider textual information into KRL. NTN \cite{socher2013reasoning} attempts to catch the potential textual relationships between entities by representing an entity using its entity name's word embeddings. \citep{wang2014knowledge_2,zhong2015aligning} propose jointly learning both entities and words embeddings by projecting them into the a unified semantic space, which aligns the entities and word embeddings using entity names, descriptions and Wikipedia anchors. And \citep{zhang2015joint,xiao2017ssp} also propose other joint frameworks for the learning of text representations and knowledge graph representations. Further, \citep{xu2016knowledge} propose to learn the models of relation extraction and knowledge graph representation jointly recently. These methods take textual information as supplements for KRL.

Another way of utilizing textual information is directly constructing knowledge graph representations from entity descriptions. Entity descriptions are often short paragraphs that provide the definitions or attributes of entities, which are maintained by some KGs or could be extracted from large datasets like Wikipedia. \citep{xie2016representation} propose DKRL which learns both entity representations based on their descriptions with CBOW or CNN encoders, and entity representations from the fact triples of KG in a unified semantic space. And the score function of DKRL is defined as:
\begin{equation}
\small
f_r(h,t)=||\mathbf{h}+\mathbf{r}-\mathbf{t}||_{L_1/L_2}+||\mathbf{h}_D+\mathbf{r}-\mathbf{t}||_{L_1/L_2}+||\mathbf{h}+\mathbf{r}-\mathbf{t}_D||_{L_1/L_2}+ ||\mathbf{h}_D+\mathbf{r}-\mathbf{t}_D||_{L_1/L_2}
\end{equation}
where $\mathbf{h}_D$ and $\mathbf{t}_D$ are the text-based representation of $h$ and $t$ which are obtained from the entity descriptions.
Note that the description-based representation could be built to represent an entity even if the entity is not in training set. Therefore, the DKRL model is capable of handling zero-shot scenario. Recently, \citep{fan2017distributed} also propose a logistic approach which also both learns entity representations based on their descriptions and learns entity representations from the fact triples of KG and achieve a better performance.

To model the complex relations in KG, \citep{Wang:2016:TRL:3060621.3060801} propose TEKE which enhances the representation of both head/tail entities and relation with the representations of its neighbor entities with similar text when models a fact triple. TEKE first calculates a co-occurrence matrix which each element $y_{ij}$ indicates co-occurrence frequency between the texts of $e_i$ and $e_j$. And then TEKE defines $n(e_i) = \{e_j|y_{ij}>\theta\}$ ($\theta$ is a hyper-parameter) as the set of neighbor entities of entity $e$, and defines $n(e_1, e_2) = n(e_1)\bigcap n(e_2)$. Hence, the representations of neighbor entities is defined as:
\begin{eqnarray}
\mathbf{n}(e_i) &=& \frac{1}{\sum_{e_j\in n(e_i)}y_{ij}}\sum_{e_j\in n(e_i)}y_{ij}\mathbf{e}_j,\\
\mathbf{n}(e_i, e_j) &=&  \frac{1}{\sum_{e_k\in n(e_i, e_j)}\min(y_{ik},y_{jk})}\sum_{e_k\in n(e_i, e_j)}\min(y_{ik},y_{jk})\mathbf{e}_k.
\end{eqnarray}
And the score function of TEKE is defined as:
\begin{equation}
f_r(h,t) = ||\mathbf{n}(h)\mathbf{M_e}+\mathbf{h}+\mathbf{n}(h, t)\mathbf{M_r}+\mathbf{r} - \mathbf{n}(t)\mathbf{M_e}-\mathbf{t}||_{L_1/L_2},
\end{equation}
where $\mathbf{M_e}$ and $\mathbf{M_r}$ are mapping matrices.


\subsubsection{Type Information} Besides textual information, entity type information, which can be viewed as a kind of label of entities, is also useful for KRL. There are some KGs such as Freebase and DBpedia possessing their own entity types. An entity could belong to multiple types, and these entity types are usually arranged with hierarchical structures. For example, \emph{William Shakespeare} have both hierarchical types \emph{book/author} and \emph{music/artist} in Freebase.

\citep{krompassISWC2015,chang-EtAl:2014:EMNLP2014} takes type information as type constraints in KRL, aiming to distinguish entities which belong to the same types. Their methods improve both performance of RESCAL \cite{chang-EtAl:2014:EMNLP2014} and TransE \cite{krompassISWC2015}. Instead of merely considering type information as type constraints, \citep{guo2015semantically} proposes semantically smooth embedding (SSE) which incorporates the type information into KRL by forcing the entities which belongs to the same type to be close to each other in the semantic space. SSE employs two kinds of learning algorithm including Laplacian eigenmaps \cite{belkin2002laplacian}:
\begin{equation}
R = \sum_{e_1\in E}\sum_{e_2\in E} g(e_1, e_2)||\mathbf{e}_1-\mathbf{e}_2||_{L_2},
\end{equation}
where $g(e_1, e_2) = 1$ if $e_1$ and $e_2$ have the same type. Or locally linear embedding \cite{roweis2000nonlinear}:
\begin{equation}
R = \sum_{e_1\in E} ||\mathbf{e}_1-\sum_{e_2\in N(e_1)} g(e_1, e_2)\mathbf{e}_2||_{L_2},
\end{equation}
where $N(e_1)$ indicates the set of the neighbors of entity $e_1$. And then $R$ is incorporated as a regularization of the overall loss function when learning the knowledge graph representation. However, SSE still has a problem that it cannot utilize the hierarchy located in the entity types.

To address this issue, \citep{hu2015entity} learn entity representations considering the whole entity hierarchy of Wikipedia. Further, TKRL \cite{xie2016representation_t} utilizes hierarchical type structures to help to learn the embeddings of entities and relations of KGs, especially for those entities and relations with few fact triples. Inspired by the idea of multiple entity representations proposed in TransR, TKRL constructs projection matrices for each hierarchical type, and the score function of TKRL is defined as follows:
\begin{equation}
\begin{split}
f_r(h,t)=||\mathbf{M}_{rh}\mathbf{h}+\mathbf{r}-\mathbf{M}_{rt}\mathbf{t}||_{L1/L2},
\end{split}
\end{equation}
where $\mathbf{M}_{rh}$ and $\mathbf{M}_{rt}$ are two projection matrices for $h$ and $t$ depending on their corresponding hierarchical types in this triple, which are constructed by hierarchical type encoders. As the head entities of a relation may have several types,  $\mathbf{M}_{rh}$ is defined as a weighted sum of the matrices of all involved hierarchical types (The same to $\mathbf{M}_{rt}$):
\begin{equation}
\mathbf{M}_{rh} = \frac{\sum_c\alpha_c\mathbf{M}_c}{\sum_c\alpha_c},
\end{equation}
where $\alpha_c = 1$ if the type $c$ is in the hierarchical type set of head entity of relation $r$, otherwise $0$. Further, the hierarchical type encoders regard sub-types as projection matrices, and utilize multiplication or weighted summation to construct projection matrices for each hierarchical type, i.e.,
\begin{eqnarray}
\text{Multiplication:}& & \mathbf{M}_c = \prod_j \mathbf{M}_{c^{(j)}},\\
\text{Weighted summation:}& & \mathbf{M}_c = \sum_j \beta_j\mathbf{M}_{c^{(j)}},
\end{eqnarray}
where $c^{(j)}$ is the $j$-th sub-type of $c$ and $\mathbf{M}_{c^{(j)}}$ is the corresponding projection matrix.

\subsubsection{Visual Information} Besides textual and type information, visual Information such as images, which can provide an intuitive outlook of their corresponding entities', is also useful for KRL. The reason is that the visual information may give significant hints suggesting some inherent attributes of entities from certain aspects. 


\citep{xie2016image} propose a novel KRL approach, Image-embodied Knowledge Representation Learning (IKRL),  to take visual information into consideration when learning representations of the KGs.  Specifically, IKRL first constructs the image representations for all entity images with neural networks, and then project these image representations from image semantic space to entity semantic space via a transform matrix. Since most entities may have multiple images with different qualities, IKRL selects the more informative and discriminative images via an attention mechanism. Finally, IKRL defines the score function following the framework of DKRL:
\begin{equation}
f_r(h,t)=||\mathbf{h}+\mathbf{r}-\mathbf{t}||_{L_1/L_2}+||\mathbf{h}_I+\mathbf{r}-\mathbf{t}||_{L_1/L_2}+||\mathbf{h}+\mathbf{r}-\mathbf{t}_I||_{L_1/L_2}+ ||\mathbf{h}_I+\mathbf{r}-\mathbf{t}_I||_{L_1/L_2}
\end{equation}
where $\mathbf{h}_I$ and $\mathbf{t}_I$ are the text-based representation of $h$ and $t$

The evaluation results of IKRL not only confirm the significance of visual information in understanding entities but also show the possibility of a joint heterogeneous semantic space. Moreover, the author also finds some interesting semantic regularities in visual space similar to  $\mathbf{v}(man)-\mathbf{v}(king) \simeq \mathbf{v}(woman)-\mathbf{v}(queen)$ found in word space.


\subsubsection{Logic Rules} Most existing KRL methods only consider the information of each relational fact separately, ignoring the interactions and correlations between different triples. Logic rules, which are usually the summaries of experience deriving from human beings' prior knowledge, could help us for knowledge reasoning. For example, if we know the triple fact that (\emph{Obama}, \texttt{president\_of}, \emph{United States}), we can easily infer with high confidence that (\emph{Obama}, \texttt{nationality}, \emph{United States}), since we know the logic rule that the relation \texttt{president\_of} $\Rightarrow$ \texttt{nationality}.

Recently, there are some works attempting to introduce logic rules to knowledge acquisition and inference. ALEPH \cite{muggleton1995inverse}, WARMR \cite{dehaspe1999discovery},  and AMIE \cite{galarraga2013amie} utilize  Markov logic networks to extract logic rules in KGs. \citep{pujara2013knowledge,beltagy2014efficient,wang2015knowledge} also utilize Markov logic networks to take the logic rules into consideration when extracting knowledge.  Besides, \citep{rocktaschel2015injecting} attempt to incorporate first-order logic domain knowledge into matrix factorization model to extract unknown relational facts from plain text. \citep{rocktaschel2014low,wang2016learning} further learn low dimensional embeddings of logic rules.

Recently, KALE \cite{guo2016jointly} incorporates logic rules into KRL via modeling the triples and rules jointly. For the triple modeling, KALE follows the translation assumption with minor alteration and the score function of KALE is defined as follows:
\begin{equation}
f_r(h, t)=1-\frac{1}{3\sqrt{d}}||\mathbf{h}+\mathbf{r}-\mathbf{t}||,
\end{equation}
where $f_r(h,t)$ takes value in $[0,1]$ for the convenience of joint learning.

To model the new-added rules, KALE employs the t-norm fuzzy logics proposed in \cite{hajek1998metamathematics}.
Specially, KALE uses two typical types of logic rules. The first one is $\forall h,t:(h,r_1,t)\Rightarrow(h,r_2,t)$ which is the same as the example above. KALE represents the scoring function of this logic rule $f_1$ as follows: 
\begin{equation}
\begin{split}
I(f_1)=f_{r_1}(h,t)\cdot f_{r_2}(h, t)-f_{r_1}(h, t)+1.
\end{split}
\end{equation}

The second logic rules is $\forall h,e,t:(h,r_1,e)\wedge(e,r_2,t)\Rightarrow(h,r_3,t)$ (e.g. given (\emph{Barbara Pierce Bush}, \texttt{father}, \emph{George W. Bush})) and (\emph{George W. Bush}, \texttt{father}, \emph{George H. W. Bush}), we can infer that (\emph{Barbara Pierce Bush}, \texttt{grandfather}, \emph{George H. W. Bush})). And KALE define the second scoring function as:
\begin{eqnarray}
I(f_2)&=&f_{r_1}(h, e)\cdot f_{r_2}(e, t)\cdot f_{r_3}(h, t)-f_{r_1}(h, e)\cdot f_{r_2}(e, t)+1.
\end{eqnarray}

The joint training strategy takes all positive formulae including fact triples as well as logic rules into consideration. In fact, the path-based TransE \cite{lin2015modeling} stated above also implicitly considers the latent logic rules between different relations via relational paths.

It is natural that we learn things in the real world with all kinds of multi-source information. Multi-source information such as plain texts, hierarchical types, or even images and videos, is of great importance when modeling the complicated world and constructing cross-modal representations. The success in these preliminary attempts demonstrates the significance and feasibility located in multi-source information, while there are still improvements to existing methods remaining to be explored. Moreover, there are still some other types of information which could also be encoded into KRL.

\section{Training Strategies}
In this section, we will introduce the training strategies for KRL models. There are two typical training strategies including margin-based approach and logistic-based approach.

\subsection{Margin-based Approach}
The margin-based approach defines the following loss function as training objective:
\begin{equation}
\label{eq:margin}
L(\theta) = \sum_{(h, r, t) \in S}\sum_{(h', r, t') \in S'}\max \big( 0, f_r(h, t) + \gamma - f_r(h', t')\big),
\end{equation}
where $\theta$ indicates all parameters of the KRL models, $\max(x,y)$ returns the higher value between $x$ and $y$, $\gamma$ is the margin and $S'$ is the set of invalid fact triples.

\textbf{Generating Invalid Triple Set}. In fact, existing KGs only contain valid fact triples, and therefore we need to generate invalid triples for the training of margin-based approach. Researchers have proposed to generate invalid triples $(h, r, t) \in S$ by randomly replacing entities or relations in valid fact triples. Hence, the invalid triple set is defined as follows:
\begin{equation}
\mathbf{S}^{-} = \bigcup_{(h, r, t)\in S} \{(h', r, t)\} \cup  \{(h, r', t)\} \cup  \{(h, r, t')\}.
\end{equation}


However, generating invalid triple set by uniformly replacement may lead to some errors. For example, the triple (\emph{Bill Gates}, \texttt{nationality}, \emph{United States}) may generate false invalid triple (\emph{Jobs Steve}, \texttt{nationality}, \emph{United States}). In fact,  \emph{Jobs Steve} is actually Americans. To alleviate this issue, when generating the invalid triple, \citep{wang2014knowledge} proposed to assign different weights for head/tail entity replacement according to the relation characteristic.  For example, for 1-to-n relation, they will tend to replace the ``one'' side instead of the ``n'' side, and therefore the probability to generate  false-invalid fact triples will be reduced. 


Besides, the uniform generating approach may not be able to generate representative negative training triples. For example,  the triple (\emph{Bill Gates}, \texttt{nationality}, \emph{United States}) may generate invalid triple (\emph{Bill Gates}, \texttt{nationality}, \emph{Jobs Steve}). In fact,  \emph{Jobs Steve} is not a nation and such negative fact triple cannot fully train the KR models. Therefore, \citep{socher2013reasoning} propose to generate negative triples by replacing entities with other entities of the same type.

\subsection{Logistic-based Approach}

The logistic-based approach defines the following loss function as training objective:

\begin{equation}
\label{eq:logistic}
L(\theta) = \sum_{(h, r, t) \in S}\log(1+ \exp(-g_r(h, t))+ \sum_{(h', r, t') \in S'}\log(1+ \exp(+g_r(h, t))
\end{equation}
where  $g_r(h,t)$ indicates that energy of the fact triple (h, r, t), which is further defined as:
\begin{equation}
g_r(h, t) = -f_r(h, t) + b, 
\end{equation}
where $b$ is a bias constant. 

And then the optimization of Eq. \ref{eq:margin} and Eq. \ref{eq:logistic} can be easily carried out by SGD \cite{bottou2010large}, Adagrad \cite{duchi2011adaptive}, Adadelta \cite{zeiler2012adadelta} or Adam \cite{kingma2014adam}.
\section{Applications of Knowledge Graph Representation}

Recent years have witnessed the great success in knowledge-driven applications such as information retrieval and question answering. These applications are expected to help accurately and deeply understand user requirements, and then appropriately give responses. Hence, they cannot work well without certain external knowledge.

However, there are still some gaps in the knowledge stored in KGs and the knowledge used in knowledge-driven applications.  To address this issue, researchers employ KRL to bridge the gap between them. Knowledge graph representations are capable of solving the data sparsity and modeling the relatedness between entities and relations. Moreover, they are convenient to be included in deep learning methods and by nature posses potential in the combination with heterogeneous information.

In this section, we will introduce typical applications of KRL including three knowledge acquisition tasks and other tasks.

\subsection{Knowledge Graph Completion}

Knowledge graph completion aims to predict the missing entities or relations  for given uncompleted fact triples. In this task, to evaluate the KRL approaches more effectively, we do not only  give a best prediction, but give a detailed ranking lists of all the entities or relations in KGs.
\subsubsection{Datasets}
In this paper, we select three typical KGs WordNet, Freebase and Wikidata to evaluate the knowledge graph representation models. For WordNet, we employ a widely-used dataset \texttt{WN18} used in \cite{bordes2014semantic} 
And for Freebase,  we also select a widely-used dataset from Freebase \texttt{FB15K} used in \cite{bordes2014semantic}. 

For FB15k, we find that there exists some direct relatedness between the fact triples between its training set and testing set, which prevents us giving a exact evaluation of various KRL approaches. The reason is that some relations such as \texttt{contains} may have its reverse relation \texttt{contained by} in testing set. Therefore, we also sample a dataset from Wikidata, named as \texttt{WD50k}, to further evaluate the performance of these KRL models. We list statistics of these data sets in Table \ref{table:statistics}.

\begin{table}[h]
\centering
\small
\caption{Statistics of datasets.}
\label{table:statistics}
\begin{tabular}{c|rrrrr}
\hline
Dataset &\#Rel& \#Ent& \#Train& \#Valid& \# Test\\
\hline
WN18  & 18	& 40,943	&141,442	& 5,000	& 5,000\\
FB15K & 1,345 & 14,951 & 483,142 & 50,000& 59,071\\
WD50K & 378    & 50,000    & 249,188    & 24,972    & 25,131\\
\hline
\end{tabular}
\end{table}

\subsubsection{Evaluation Results}

As set up in \cite{bordes2013translating}, we adopt the following evaluation metrics: (1) Mean Rank, which indicates the mean rank of all correct predictions; and (2) Hits@10, which is the proportion of correct predictions ranked in top-$10$.  We also use two settings ``Raw'' and ``Filter'', where the ``Filter'' setting will filter out the other correct entities when measuring evaluation metrics.

\begin{table*}[!h]
  \centering
  \scriptsize
  \caption{Evaluation results on link prediction.}
  \label{label:link_prediction}
  \begin{tabular}{c|rr|rr|rr|rr|rr|rr}
    \hline
    Data Sets & \multicolumn{4}{|c|}{WN18}&\multicolumn{4}{|c}{FB15K}&\multicolumn{4}{|c}{WD50K}\\
    \hline
    \multirow{2}{*}{Metric} & \multicolumn{2}{|c|}{Mean Rank} & \multicolumn{2}{|c|}{Hits@10 ($\%$)} & \multicolumn{2}{|c|}{Mean Rank}& \multicolumn{2}{|c}{Hits@10 ($\%$)} & \multicolumn{2}{|c|}{Mean Rank}& \multicolumn{2}{|c}{Hits@10 ($\%$)}\\
    & Raw &Filter & Raw & Filter & Raw & Filter & Raw & Filter  & Raw & Filter & Raw & Filter\\
    \hline
    \multicolumn{13}{c}{\emph{Linear models}}\\
    \hline
    RESCAL        & 1,180  & 1,163 & 37.2 & 52.8 &    828 & 683 &  28.4 & 44.1 & - & - & - & -\\
    SE            & 1,011  &    985 & 68.5 & 80.5 &    273 & 162 &  28.8 & 39.8 & - & - & - & -\\
    SME (linear)        &    545  &    533 & 65.1 & 74.1 &    274 & 154 &  30.7 & 40.8 & - & - & - & -\\
    SME (bilinear)    &    526  &    509 & 54.7 & 61.3 &    284 & 158 &  31.3 & 41.3 & - & - & - & -\\
    LFM            &    469  &   456 & 71.4 & 81.6 &    283 & 164 &  26.0 & 33.1 & - & - & - & -\\
    \hline
    \multicolumn{13}{c}{\emph{Translation models}}\\
    \hline
    TransE            & 263    & 251    & 75.4    & 89.2    & 246    & 92        & 48.8    & 74.5    & 787    & 548    & 37.0    & 48.4\\
    TransH            & 404    & 391    & 78.0    & 90.3    & 230    & 78        & 50.7    & 76.7    & 686    & 438    & 39.4    & 52.0\\
    TransR          & 423    & 410    & 80.4    & 93.8    & 230    & 79        & 51.8     & 79.0    & 1015    & 753    & 39.6    & 52.8\\
    TransD          & 415       & 401       & 80.3    & 93.5    & 236    & 87        & 51.2    & 77.5    & 924    & 668    & 38.4    & 50.0\\
    TranSparse        & 383    & 369    & 80.2    & 93.1    & 256    & 99        & 48.9    & 75.9    & 730    & 475    & 38.3    & 50.7\\
    PTransE        & -        & -        & -        & -        & 207    & 58        & 51.4    & 84.6& - & - & - & -\\
    \hline
    \multicolumn{13}{c}{\emph{Other models}}\\
    \hline
    HolE             & 730    & 710    & 82.5    & 94.3    & 442    & 288    & 44.7    & 70.4			& 4614 & 4353 & 24.3 & 35.9\\
    ComplEx        & 859    & 844    & 80.1    & 94.0    & 261    & 97        & 48.3    & 84.0		& 1510 & 1245 & 35.4 & 47.6\\
    \hline
  \end{tabular}
\end{table*}

In this section, we discuss the performance in detail to gain more insights about what really works for KRL. Evaluation results on WN18, FB15K and WN50k are shown in Table \ref{label:link_prediction}. From the table, we can see that:

(1) All the models with complex relation modeling including TransH, TransR, TransD, TranSparse, and KG2E outperform TransE in Hits@10 and Mean Rank on both datasets significantly. The reason is that TransE cannot deal with the complex relations in KG but these models attempt to alleviate the issue. 

(2) By taking the relational path into consideration, PTransE achieves the best performance among all models on FB15k. It indicates that there exist complex relation inferences in KG and it can benefit the KRL.


(3) On WN18, we find that for all the models, when the dimension $d$ arises, the performance in Hits@10 will improve when the performance in Mean Rank will decrease. The reason is perhaps that the increase of dimension d could improve the discrimination of entities and relations especially for those entities with a large number of fact triples. However, for those entities with a few fact triples, the increase in dimension $d$ may lead to insufficient learning which may influence the system performance.

\subsubsection{Analysis}

Translation models, HolE and ComplEx have achieved promising results in the task of knowledge graph completion. To conduct an in-depth analysis of these models, we select and re-implement eight typical models including TransE, TransH, TransR, TransD, TranSparse, PTransE, HolE and ComplEx.  In this section, we compare the performance of the selected models in different mapping properties, dimensions, and margins.

{
\begin{table*}[htb]
\centering
\footnotesize
\caption{Evaluation results on FB15K by mapping properties of relations. ($\%$)}
 \label{label:mapping_property}
\begin{tabular}{c|rrrr|rrrr}
    \hline
    Tasks &\multicolumn{4}{c|}{Predicting Head (Hits@10)}&\multicolumn{4}{|c}{Predicting Tail (Hits@10)}\\
    \hline
    Relation Category&1-to-1&1-to-N&N-to-1&N-to-N&1-to-1&1-to-N&N-to-1&N-to-N\\
    \hline
    \multicolumn{9}{c}{\emph{Translation models}}\\
    \hline
    TransE        & 68.7       & 90.7     & 37.7     & 76.3     & 67.6     & 48.3     & 90.4     & 78.5\\
    
    TransH        & 79.8    & 92.7    & 42.7    & 77.5    & 78.8    & 53.2    & 92.0    & 80.4\\
    TransR        & 84.6    & 94.9    & 48.2    & 79.4    & 84.3    & 57.3    & 93.7    & 82.6\\
    TransD        & 79.9    & 92.4    & 43.3    & 77.7    & 80.4    & 52.9    & 92.7    & 80.7\\
    
    TranSparse    & 79.5    & 90.1    & 38.1    & 78.4    & 78.4    & 49.7    & 90.5    & 80.3\\
    PTransE     & 90.1    & 92.0    & 58.7    & 86.1    & 90.7    & 70.7    & 87.5    & 88.7\\
    \hline
    \multicolumn{9}{c}{\emph{Other models}}\\
    \hline
    HolE        & 76.3    & 65.1    & 40.9    & 75.2    & 75.2    & 53.3    & 53.7    & 77.2\\
    ComplEx    & 80.4    & 88.9    & 57.1    & 84.9    & 80.8    & 66.7    & 80.7    & 86.1\\
    \hline
\end{tabular}
\end{table*}
}

We categorize the relations according to their characteristics into four classes:1-to-1, 1-to-n, n-to-1, n-to-n. In Table \ref{label:mapping_property}, we show separate evaluation results of these four types of relations on FB15K. We can observe that:

 (1) All TransE's extensions considering complex relation modeling achieve better results for the  ``1-to-n'', ``n-to-1'' and ``n-to-n'' relations as compared to TransE. It indicates that these models actually improve the ability to model complex relations.
 
 (2) PTransE also performs better for the ``1-to-1" relations as compared to TransE. It indicates that these models obtain better representations of entities and relations by especially dealing with complex relations. 
 
 (3) PTransE achieves the best performance among all models in all mapping properties. It indicates that PTransE obtain better representations of entities and relations by taking the relational paths into consideration and relation inference can benefit to knowledge graph completion.
 
 (4) ComplEx performs better as compared to all translation models except PTransE which considers the information of relational paths. It demonstrates that the complex embeddings are more suitable to represent KGs as compared to traditional real vectors.

For all above models, there are two hyper-parameters which have a significant influence on the performance: the dimension $d$ and the margin $\gamma$. Hence, we further compare the performance with respect to these two hyper-parameters on the dataset FB15k in Hits@10. For other hyper-parameters, we use the same setting as the task of knowledge graph completion on FB15k.

\textbf{Effect of dimension d.}

\begin{table*}[htb]
\centering
\footnotesize
\caption{Parameter sensitivity of dimension d (Hits@10)}
\label{label:dim}
    \begin{tabular}{c|c|c|c|c|c|c|c|c|c|c}
        \hline
        Data Sets &\multicolumn{5}{|c}{FB15K}&\multicolumn{5}{|c}{WD50K}\\
    \hline
    Dimension    & 25        & 50        & 100    & 200    & 400    & 25        & 50        & 100    & 200    & 400    \\
    \hline
    TransE        & 49.5    & 60.0    & 68.8    & 74.4    & 74.5    & 42.0    & 46.7    & 47.0    & 48.4    & 48.0    \\
    TransH        & 51.6    & 60.9    & 70.7    & 76.7    & 76.6    & 42.2    & 47.3    & 49.2    & 52.0    & 48.1    \\
    TransR        & 56.7    & 67.2    & 75.5    & 79.0    & 74.5    & 41.5    & 48.2    & 52.2    & 52.8    & 52.1    \\
    TransD        & 51.1    & 61.4    & 72.1    & 77.5    & 76.9    & 41.7    & 46.5    & 48.8    & 50.0    & 49.0    \\
    TranSparse & 51.1    & 62.2    & 71.8    & 74.2    & 75.9    & 38.4    & 46.2    & 48.9    & 50.7    & 47.2    \\
    HolE            & 37.5    & 57.0    & 69.4    & 70.4    & 62.4    & 34.4    & 35.9    & 32.4   &  28.1    & 20.7   \\
    ComplEx     & 70.1    & 82.4    & 82.8    & 83.4    & 84.0    & 22.3    & 34.0    & 38.5    & 42.3    & 47.6     \\
    \hline
    \end{tabular}
\end{table*}

 Evaluation results are shown in Table \ref{label:dim}. From the table, we observe that:

(1) All the models achieve better performance in dimension $100$, $200$ and $400$, and the system performance doesn't improve significantly when the dimension is greater than 400.

(2) ComplEx is more robust as compared to all other models even for TransR with much more parameters, which indicates that the complex embeddings make the model more expressive.

\textbf{Effect of margin.} 
\begin{table*}[!h]
\centering
\footnotesize
\caption{Parameter sensitivity of margin (Hits@10)}
\label{label:margin}
    \begin{tabular}{c|c|c|c|c|c|c|c|c|c|c}
    \hline
        Data Sets &\multicolumn{5}{|c}{FB15K}&\multicolumn{5}{|c}{WD50K}\\
    \hline
    Margin        & 0.5        & 1        & 2        & 4       &	8 & 0.5        & 1        & 2        & 4	& 8\\
    \hline
    TransE        & 68.3    & 68.8    & 67.2    & 58.3  & 52.1  & 43.5    & 46.4    & 47.1    & 48.4 & 48.0\\
    TransH        & 70.4    & 70.9    & 68.6    & 61.5  & 53.2  & 46.8    & 48.9    & 50.8    & 52.0 & 49.7\\
    TransR        & 74.5    & 75.3    & 74.9    & 68.2  & 55.4   & 51.8    & 52.8     & 51.8    & 49.9 & 49.9\\
    TransD         & 67.4    & 72.2    & 69.2    & 59.3  &53.6  & 46.1    & 49.2    & 50.0    & 49.6 & 49.4\\
    TranSparse    & 70.6    & 71.8    & 70.0    & 45.3  & - & 45.8    & 47.9    & 48.7    & 50.7 & -\\
    \hline

     \end{tabular}
\end{table*}

Evaluation results are shown in Table \ref{label:margin} (As HolE and ComplEx don't have this hyper-parameter, we don't list their results here). From the table, we observe that:

(1) All the models perform well when the margin $\gamma = 0.5/1.0/2.0$. Therefore these models can keep stable when the margin within a reasonable range.

(2) All the models cannot perform well when the margin $\gamma = 4.0$. But TransR performs better as compared to other models when the margin $\gamma = 4.0$. The reason is perhaps that TransR has much more parameters than other models and its strong model ability makes it more robust.

\subsubsection{Type Constraints}

In addition to the fact triples, most existing KGs such as Wikidata also provide type-constraints information for relations which gives the type constraints of the head and tail entities for each relation. The prior knowledge of relations provides additional information for KRL, e.g. that the relation \texttt{nationality} should relate only head entity of the type \emph{Person} and tail entity of the type \emph{Country}.

It has been proved that take the type-constraints information of relation into account could help KRL approaches to model entities and relations in KG\cite{krompassISWC2015}. We also report the Hit@10 for all models with type constraints (+TC) in Table \ref{label:link_prediction_type}.

\begin{table}[htb]
  \centering
\footnotesize
  \caption{Evaluation results on link prediction with type constraints.}
  \label{label:link_prediction_type}
  \begin{tabular}{c|c|l|c|c}
    \hline
    \multirow{2}{*}{Models}        & \multicolumn{2}{|c|}{FB15k}    &\multicolumn{2}{|c}{WD50k} \\
    \cline{2-5}
                            & Origin         & +TC         & \multicolumn{1}{|c|}{Origin}        & \multicolumn{1}{|c}{+TC}\\
    \hline
    TransE                    & 74.5        & 78.7 (+4.2)    & 48.4    & 49.9 (+1.5) \\
    TransH                    & 76.7        & 80.0 (+3.3)    & 52.0    & 53.6 (+1.6) \\
    TransR                    & 79.0        & 81.9 (+2.9)    & 52.8    & 54.5 (+1.7) \\    
    TransD                    & 77.5        & 80.0 (+2.5)    & 50.0    & 51.4 (+1.4) \\    
    TranSparse                & 75.9        & 79.8 (+3.9)    & 50.7    & 52.3 (+1.6) \\
    HolE					  & 70.6		& 81.6 (+11.0)   &	35.9   & 45.6(+9.7)		\\
    ComplEx		& 84.0	   & 87.2	(+3.2)	& 47.6		&	54.1 (+6.5)	\\
    \hline
  \end{tabular}
\end{table}

From the table we can see that: All the models have shown great improvement when considering type constraints. It indicates that the type-constraints information of relations provided by the KGs are useful for existing KRL methods in modeling KGs and further knowledge driven tasks.

\subsection{Triple Classification}

Triple classification aims to distinguish if  a given triple is correct or not, which has been studied in \cite{socher2013reasoning,wang2014knowledge} as one of their evaluation tasks. Here, we use three typical datasets in this task including WN11, FB13 and FB15K, where the first two datasets are used in \cite{socher2013reasoning} and their statistics are listed in Table \ref{table:statistics-tc}.

\begin{table}[h!]
\centering
\small
\caption{Statistics of datasets.}
\label{table:statistics-tc}
\begin{tabular}{c|rrrrr}
\hline
Dataset &\#Rel& \#Ent& \#Train& \#Valid& \# Test\\
\hline
WN11  &11       & 38,696 & 112,581 & 2,609  &10,544\\
FB13   &13       & 75,043 &316,232  &5,908  &23,733\\
\hline
\end{tabular}
\end{table}


\begin{table}[htb]
\centering
\footnotesize
\caption{Evaluation results of triple classification. ($\%$)}
\label{label:triple_classification}
\begin{tabular}{c|r|r|r}
\hline
Data Sets & WN11 & FB13 & FB15K \\
\hline
SE                       & 53.0     & 75.2    & - \\
SME (bilinear)    & 70.0    & 63.7    & - \\
SLM                   & 69.9    & 85.3    & - \\
LFM                    & 73.8    & 84.3    & - \\
NTN                    & 70.4    & 87.1    & 68.5 \\
\hline
TransE        & 85.0        & 83.1        & 79.6    \\
TransH            & 85.5        & 83.7        & 80.2    \\
TransR            & 85.2        & 82.5        & 83.9    \\
TransD            & 85.6        & 81.4        & 81.4    \\
TranSparse        & 85.6        & 84.0        & 84.2    \\
HolE            & -        & -        & 85.9    \\
ComplEx        & -        & -        & 87.2    \\
\hline
\end{tabular}
\end{table}

The experimental results of triple classification is shown in Table \ref{label:triple_classification}. From the table, we have the following observations: 

(1) On WN11, TransE and its extension have similar performance. The reason is perhaps that WN11 only has 11 relationships which are too simple to distinguish the model ability of different translation models. 

2) None of TransE and its extensions can outperform NTN on FB13 with only $13$ relations. In contrast, on the more sparse dataset FB15K with $1,345$ relations, TransE and its extensions have much better performance as compared to NTN. The reason is perhaps that NTN is more expressive while maintains much more parameters.  Therefore, it performs better in the dense graphs, while suffers from the lack of data in sparse graphs. On the contrary,  TransE and its extensions are more simple and effective,  achieving promising result in sparse graphs.

\subsection{Relation Extraction}
Relation extraction (RE) aims to extract unknown relational fact from plain text on the web, which is an important information source to enrich KGs. Recent, distantly supervised RE models \cite{mintz2009distant,riedel2010modeling,hoffmann2011knowledge,surdeanu2012multi} have become the mainstream approaches to extract novel facts from plain texts. However,  these methods only use the information in  plain text in knowledge acquisition, ignoring the rich information contained by the structure of KGs. 

\cite{weston2013connecting} proposes to combine TransE and existing distantly supervised RE models to extract novel facts, and obtains lots of improvements. Moreover, \citep{han2016joint} propose a novel joint representation learning framework for KRL and RE. In this section, we will investigate if existing KRL models could effectively enhance existing distantly supervised RE models.

Following \cite{weston2013connecting}, we adopt a widely used dataset NYT10 which is developed by \cite{riedel2010modeling} in our experiments. This dataset contains $53$ relations, and $18,252$ relational facts as well as $1,950$ relational facts in training and testing sets respectively. Besides, the training  and testing set contain $522,611$ and $172,448$ sentences respectively. 

In our experiments, with loss of generality, we follow the experimental settings in \cite{lin2015learning} to implement the distantly supervised RE model named as Sm2r following\cite{weston2013connecting}, and the KRL models are all trained in  FB40k dataset which contains $39,528$ entities and $1,336$.

We combine the output scores both from Sm2r with the scores from various KRL models to predict novel relational facts, and get precision-recall curves for the models combined with TransE, TransH, TransR and PTransE.
\begin{figure}[htb]
\centering
\includegraphics[width=0.5\columnwidth]{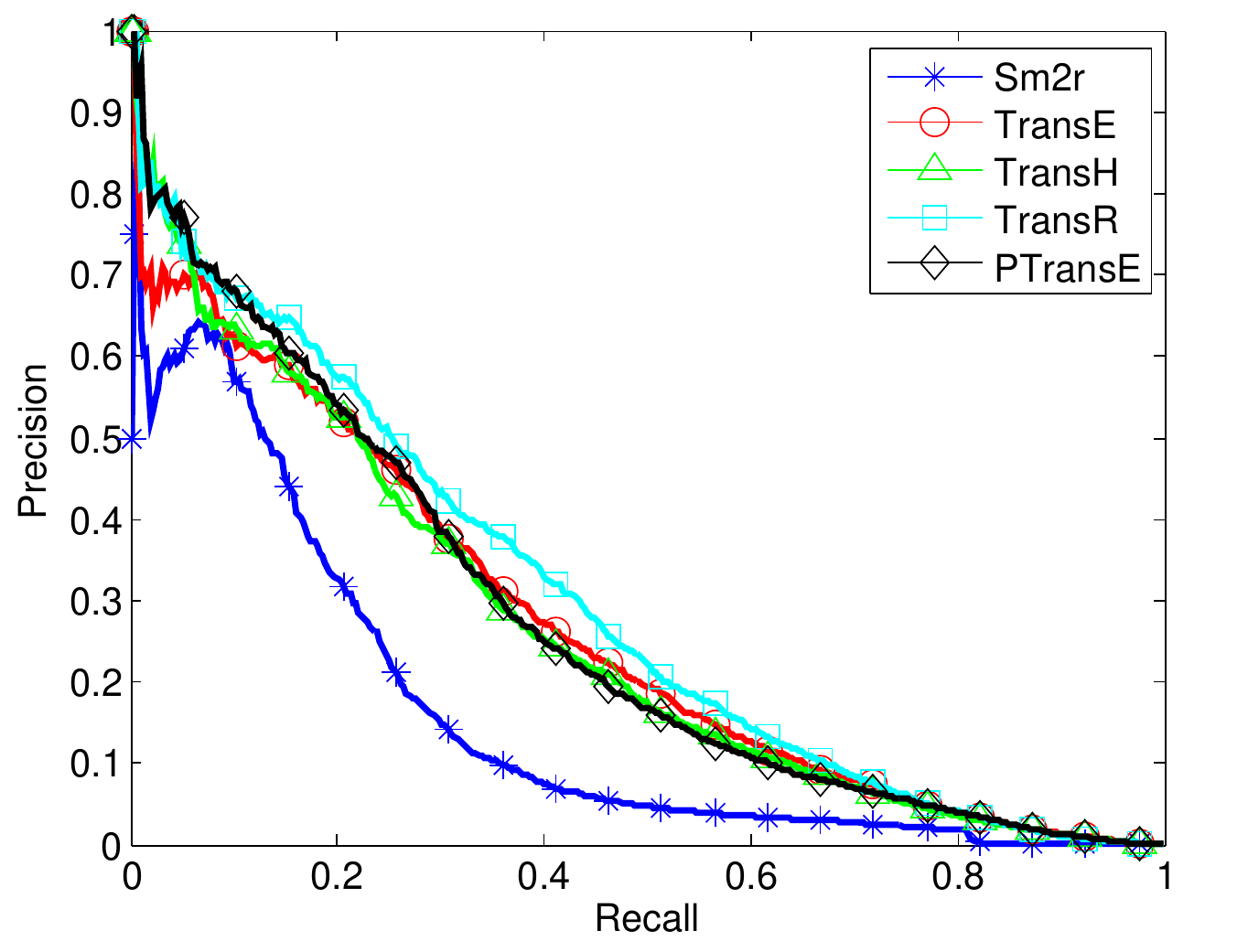}
\caption{Precision-recall curves of incorporating TransE, TransH, TransR and PTransE in distantly supervised RE.}
\label{fig:relation_extraction}
\end{figure}

From the figure, we observe that by combining existing KRL models, the performance of distantly supervised RE is much better than the original ones. It indicates that incorporating the information from KGs is useful for  distantly supervised RE.

\subsection{Other Applications}

Besides knowledge acquisition, KRL has also been applied in many other NLP task. In this section, we will introduce some typical knowledge-driven tasks including language modeling, question answering, information retrieval, recommendation system, and etc.

\subsubsection{Language Modeling} Language models aim to learn a probability distribution over sequences of words, which is a classical and essential NLP task. Recently, neural models such as RNN have proved to be effective in language modeling. However, most existing neural language models suffer from the incapability of modeling and utilizing background knowledge. The reason is that the statistical co-occurrences cannot instruct the generation of all kinds of knowledge, especially for those entities with low frequencies in plain text.

To address this issue, \citep{ahn2016neural} propose a Neural Knowledge Language Model (NKLM) that considers background knowledge provided by KGs when generating natural language sequences with RNN language models. The key is NKLM's two heterogeneous ways to generate words. One is to generate a word from the ``word vocabulary'' according to the word probabilities calculated by RNN language model, and another one is to generate a word from the ``knowledge vocabulary" according to the external KGs.

The NKLM model explores a novel neural model that combines the symbolic knowledge information in external KGs with RNN language models. However, the topic knowledge is needed when generating natural languages, which makes NKLM less practical and scalable for more general topic-independent texts. Nevertheless, we still believe that it is promising to encode knowledge into language model with such methods.

\subsubsection{Question Answering}


Question answering aims to give answers according to users' questions, which needs the capabilities of both natural language understanding on questions and inference on answer selection. Therefore, combining knowledge with question answering is a straightforward application for knowledge representations. Conventional question answering systems directly utilize KGs as background databases. These systems usually transform user's questions into regular queries and search KG for appropriate answers. However, they always ignore the potential relationships between entities and relations. Recently, with the development of deep learning, explorations have been done on neural network models for understanding questions and even generating answers.

Considering the flexibility and diversity of answer generation in natural languages, \citep{yin2016neural} propose a neural generative question answering model which explores how to utilize the facts in KGs to answer simple factoid questions. Besides, KRL models is also applied in \cite{serban-EtAl:2016:P16-1}  which attempts to generate factoid questions. Moreover, \citep{hegenerating} propose an end-to-end question answering system which incorporates copying and retrieving mechanisms to generate natural answers using KRL technique.

\subsubsection{Information Retrieval}
Information retrieval aims to retrieve related articles according to user's queries. Similar to question answering, how to exactly understand users' meanings is a crucial problem of information retrieval. Hence, incorporating the information of KG could be beneficial to information retrieval. Traditional information retrieval systems always regard user's query and retrieved articles as strings and measure their similarity using human designed feature such as bag-of-words. However, these system cannot actually realize users' meaning via simple string matching. 

Recently, with the success of KRL in many other NLP tasks, researchers have focused on utilizing KRL techniques for information retrieval. They usually improve the word-based representation used in information retrieval by entity-based representation learned by KRL methods. \citep{hasibi2015entity} propose an entity-based language model to understand users' queries in information retrieval, which is combined with word-based retrieval model to further improve the retrieval performance. Similarly, \citep{xiong2016bag} propose a bag-of-entity model which represents queries and articles with their entities. Moreover, \citep{nguyen2016toward} propose to incorporate KGs in deep neural approaches for document ranking and \citep{xiong2017explicit} represents queries and articles in the entity space, and utilize KRL to capture their semantic relatedness in KGs.

\subsubsection{Recommendation System}

With the rapid growth of web information, recommender systems have been playing an important role in web application. Recommender system aims to predict the "rating" or "preference" that users may give to items. And since KGs can provide rich information including both structured and unstructured data, recommender systems have utilized more and more KG to enrich their contexts.

\citep{cheekula2015entity} explore how to utilize the hierarchical knowledge from the DBpedia category structure in recommendation system and employ the spreading activation algorithm to identify entities of interest to the user.  Besides, Passant \cite{passant2010dbrec} measures the semantic relatedness of the artist entity in a KG to build music recommendation systems. However, most of these systems mainly investigate the problem by leveraging the structure of KGs. Recently, with the development of representation learning, \citep{zhang2016collaborative} propose to jointly learn the representations of entities in both collaborative filtering recommendation systems and KGs.

Except for the task stated above, there are gradually more efforts focusing on encoding knowledge graph representations into other tasks such as dialogue system \cite{le2016lstm,zhu2017flexible}, entity disambiguation \cite{huang2015leveraging,fang2016entity}, entity typing \cite{Ren:2016:LNR:2939672.2939822,neelakantan-chang:2015:NAACL-HLT}, knowledge graph alignment \cite{ijcai2017-209,ijcai2017-595}, dependency parsing \cite{kim2015re}, etc. Moreover, the idea of KRL has also motivated the research of visual relation extraction \cite{Zhang_2017_CVPR,baier2017improving} and social relation extraction \cite{ijcai2017399}.


\section{Discussion and Outlook}

KGs represent both entities and their relations in the form of relational triples, which provides an effective way for human beings to learn and understand the real world. Now, as a useful and convenient tool to deal with the large-scale KGs, KRL is widely explored and utilized in multiple knowledge-driven tasks, which significantly improves their performances.
	
Although existing KR models have already shown their powers in modeling KGs, there are still many possible improvements of them to be explored of. In this section, we will discuss the challenges of KRL and its applications. 

\subsection{Further Exploration of Internal and  External Information}

Relational triples, which are regarded as the internal information of KGs, have been well organized by existing KRL methods. However, the performances of these models are still far from being practical in real-world application such as knowledge graph completion.
In fact, entities and relations in KGs have their complex characteristics and rich information which have not been taken into full consideration. In this section, we will discuss the internal and external information to be further explored to enhance the performance of KRL methods.

\subsubsection{Type of Knowledge} Researchers usually divide the relations in KGs into four types including 1-to-1, 1-to-n, n-to-1 and n-to-n relations according to their mapping properties. And different KRL methods have different performance when dealing with four kinds of relations. It indicates that we need to specially design different KRL framework for different kinds of knowledge or relations. However, existing KRL methods simply divide all relations into 1-to-1, 1-to-n, n-to-1 and n-to-n relations, which cannot effectively describe the characteristics of knowledge. According to the cognitive and computational characteristics of knowledge, existing knowledge could divide into several types: (1) Hyponymy  (e.g. \texttt{has\_part})  which indicates the subordination between entities. (2)  Attribute (e.g. \texttt{nationality}) which indicates the attribute information of entities. Lots of entities may share the same attributes, especially for those enumerative attributes such as gender, age, etc. (3) Interrelation (e.g. \texttt{friend\_of}) which indicates relationships between entities. It is intuitive that these different kinds of relations should be modeled in different ways. 

\subsubsection{Dynamics of Knowledge} Existing KRL methods usually simply embed the whole KG into a unified semantic space via learning from all fact triples, neglecting the time information contained in KG. In fact, knowledge is not static and will change over time. For any point of time, there should be a unique KG with the corresponding timestamp. For instance, \emph{George W. Bush} was the president of \emph{United States} during $1995-2000$, and should not be regarded as a politician in recent years. Considering the time information of fact triples will help to understand entities and their relations more precisely in KRL. What's more, the research on the development of KGs have impacts not only on KG theories and applications but also on the study of human histories and cognition. There are some existing works \cite{jiang2016encoding,esteban2016predicting,trivedi2017know} attempting to incorporate temporal information into KRL, but their efforts are still preliminary and the dynamics of knowledge still needs to be further explored.

\subsubsection{Multi-lingual Representation Learning} \citep{mikolov2013distributed} observe a strong similarity of the geometric arrangements of corresponding concepts between the vector spaces of different languages, and suggest that a cross-lingual mapping between the two vector spaces is technically plausible. And the joint-space models for cross-lingual word representations are desirable, as language-invariant semantic features can be generalized to make it easy to transfer models across languages. Besides, there are many projects, such as DBpedia, YAGO, Freebase and so on, are constructing multilingual KGs by extracting structured information from Wikipedia. Multilingual KGs are important for the globalization of knowledge sharing and play important roles in many applications such as cross-lingual information retrieval, machine translation, and question answering.  However, to the best of our knowledge, little works have been done for representation learning of multilingual KGs. Therefore, multi-lingual KRL, which aims to improve the performances of comparative sparse KGs in some languages with the help of those of rich languages, is also a significative but challenging work to be solved. 

\subsubsection{Multi-source Information Learning} With the fast development of high-speed network, billions of people from all over the world can easily upload and share multimedia contents instantly. As what we are witnessing, not only does Internet contain pages and hyper-links nowadays. It turns out that audio, photos, and videos have also become more and more on the Web. How to efficiently and effectively utilize the multi-source information from text to video is becoming a critical and challenging problem in KRL. And multi-source information learning has shown its potential to help model KGs while existing methods of utilizing such information are still preliminary. We could design more effective and elegant models to utilize these kinds of information better. Moreover, other forms of multi-source information such as social networks are still isolated from the construction of knowledge graph representations, which could be further explored. 

\subsubsection{One-shot/Zero-shot Learning} Recently, one-shot/zero-shot learning is blooming in various fields such as word representation, sentiment classification, machine translation and so on. One-shot/zero-shot learning aims to learn from instances of an unseen class or a class with only a few instances.In the representation of KGs, the practical problem is that the low-frequency entities and relations are learned more poorly than those of high-frequency.  The representations of these low-frequent entities and relations are one of the key points to apply KGs in the real-world applications. It is natural that external information such as multi-lingual and multi-modal information can help to construct knowledge graph representations, especially for the large-scale sparse KG. We believe that with the help of multi-lingual and multi-modal representations of entities and relations, the representations of low-frequency entities and relations could be better in some degree. Besides, it's necessary to design a new KRL framework which is more suitable for the representation learning of low-frequency entities and relations.

\subsection{Complexity in Real-world Knowledge Applications}

KGs are playing an important role in a variety of applications such as web search, knowledge inference, and question answering. However, due to the complexities of real-world knowledge applications, it is still difficult to effectively and efficiently utilize KGs. In this section, we will discuss the issues which we are confronted with when utilizing KGs in real-world application.

\subsubsection{Low Quality of KGs}

One of the main challenges in real-world knowledge applications is the quality of huge KGs themselves. Typical KGs such as Freebase, DBpedia, Yago, Wikidata and so on often obtain their fact triples by automatically knowledge acquisition from huge size of plain texts on the Internet. Therefore, these KGs inevitably suffer from the issues of noise and contradiction due to the lack of human labeling. These noises and conflicts will lead to error propagation when involves with real-world application. How to automatically detect the conflict or errors in existing KGs becomes an important problem when incorporating the information of KGs into real-world application.

\subsubsection{Large Volume of KGs} The existing KGs are too cumbersome to deploy in real-world applications efficiently. They have already included millions of entities and billions of their facts about the world. For example, Freebase has $23$ million entities and $1.9$ billion triples of facts up to now. Due to huge sizes of KGs, some existing methods will be not practical because of their model and computational complexity. To the best of our knowledge, there are still many possible improvements on existing methods for leveraging both effectiveness and efficiency on the astonishing huge-size KGs.

\subsubsection{Endless Changing of KGs} Knowledge changes with time, and there are  new knowledge comes into being with time goes by. Existing KRL methods have to re-learn their models from scratch every time when the KG changes since their optimization objective is related to all the fact triples in KGs.  It is time-consuming and not practical if we want to utilize KGs in real-world application. Therefore, to design a new framework of KRL which can carry out online learning and update the model parameters incrementally is crucial to the applications of KGs.
\section{Conclusion}

In this paper, we first give a broad overview of existing approaches based on KRL, with a particular focus on three main challenges including complex relation modeling, relational path modeling, and multi-source information learning.  Secondly, we present a quantitative analysis of recent KR models and explore which factors benefit the modeling indeed in three knowledge acquisition tasks.  Thirdly, we introduce typical applications of KRL including language modeling, question answering,  information retrieval, recommendation system, etc. Finally, we discuss the remaining challenges of KRL and its application, and then give an outlook of the future study of KRL.



\bibliography{my}

\end{document}